\documentclass[10pt,twocolumn,letterpaper]{article}

\usepackage{cvpr}              
\usepackage{float}
\usepackage{pifont}
\usepackage{array}
\usepackage{color}
\usepackage{xcolor}
\usepackage{amssymb}
\usepackage{longtable}
\usepackage{multirow}
\usepackage{indentfirst}
\usepackage{caption}
\usepackage{subcaption}
\usepackage{supertabular}
\usepackage{makecell}
\usepackage{colortbl}
\usepackage{graphicx} 
\usepackage[most]{tcolorbox}
\usepackage{verbatim}

\definecolor{cvprblue}{rgb}{0.21,0.49,0.74}
\usepackage[pagebackref,breaklinks,colorlinks,allcolors=cvprblue]{hyperref}

\def\paperID{15619} 
\def\confName{CVPR}
\def\confYear{2025}

\title{VLABench: A Large-Scale Benchmark for Language-Conditioned Robotics Manipulation with Long-Horizon Reasoning Tasks}

\author{
Shiduo Zhang$^{1}$, Zhe Xu$^{1}$, Peiju Liu$^{1}$\thanks{Equal contribution}, Xiaopeng Yu$^{1}$\footnotemark[1], Yuan Li$^{1}$, Qinghui Gao$^{1}$,\\
Zhaoye Fei$^{1}$, Zhangyue Yin$^{1}$, Zuxuan Wu$^{1}$, Yu-Gang Jiang$^{1}$, Xipeng Qiu$^{1}$\thanks{Corresponding Author:sdzhang23@m.fudan.edu.cn, xpqiu@fudan.edu.cn}\\
\small $^1$School of Computer Science, Fudan University\\
\small\textsuperscript{*}Equal contribution\\
\small\texttt{sdzhang23@m.fudan.edu.cn, xpqiu@fudan.edu.cn}\\
\small\texttt{Project Website: \href{https://vlabench.github.io/}{https://vlabench.github.io/}}
}

\begin{document}

\twocolumn[{
\maketitle 
\footnotetext[1]{Corresponding author:}
\vspace{-1cm}
\renewcommand\twocolumn[1][]{#1}%
\begin{center}
\centering
\includegraphics[width=\textwidth]{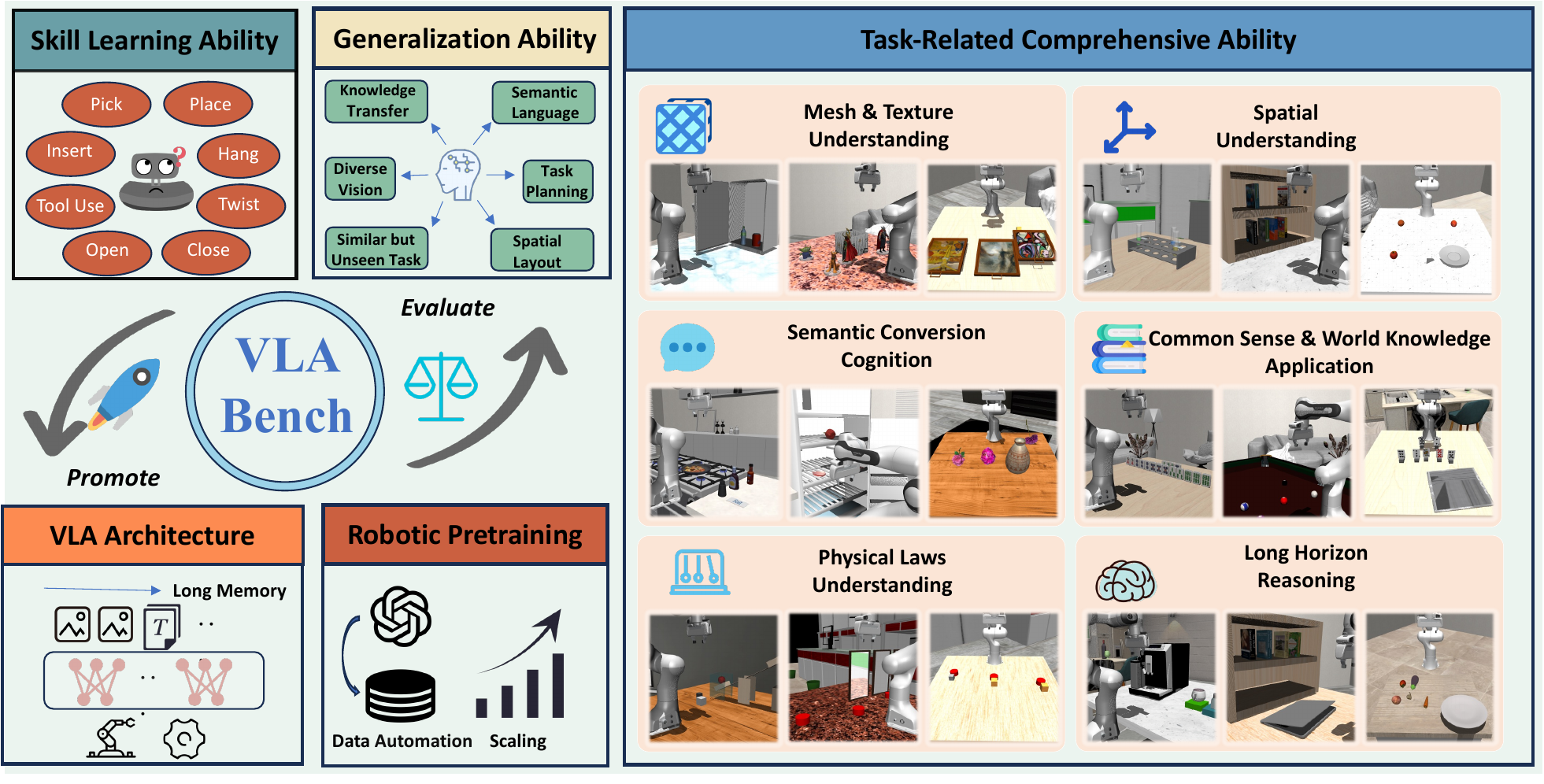}
\captionof{figure}{\textbf{Overview of VLABench.} VLABench is a large-scale language-conditioned manipulation benchmark to evaluate the comprehensive skill learning and generalization ability of action policies especially pre-trained vision-language-action models. }
\label{Fig:warping}
\end{center}
}]


\begin{abstract}
General-purposed embodied agents are designed to understand the users' natural instructions or intentions and act precisely to complete universal tasks. Recently, methods based on foundation models especially Vision-Language-Action models (VLAs) have shown a substantial potential to solve language-conditioned manipulation (LCM) tasks well. However, existing benchmarks do not adequately meet the needs of VLAs and relative algorithms. To better define such general-purpose tasks in the context of LLMs and advance the research in VLAs, we present VLABench, an open-source benchmark for evaluating universal LCM task learning. VLABench provides 100 carefully designed categories of tasks, with strong randomization in each category of task and a total of 2000+ objects. VLABench stands out from previous benchmarks in four key aspects: 1) tasks requiring world knowledge and common sense transfer, 2) natural language instructions with implicit human intentions rather than templates, 3) long-horizon tasks demanding multi-step reasoning, and 4) evaluation of both action policies and language model capabilities. The benchmark assesses multiple competencies including understanding of mesh\&texture, spatial relationship, semantic instruction, physical laws, knowledge transfer and reasoning, etc. To support the downstream finetuning, we provide high-quality training data collected via an automated framework incorporating heuristic skills and prior information. The experimental results indicate that both the current state-of-the-art pretrained VLAs and the workflow based on VLMs face challenges in our tasks. 
\end{abstract}    
\section{Introduction}
\label{sec:intro}


Language-conditioned manipulation represents a fundamental challenge in embodied AI and a stepping stone toward Artificial General Intelligence \cite{rt22023arxiv, driess2023palm, achiam2023gpt}.
Such tasks require agents to master multiple capabilities: interpreting natural language instructions, understanding complex environments, making decisions, formulating plans, and executing precise actions.
The rapid advancement of Large Language Models (LLMs) and Vision-Language Models (VLMs)~\citep{achiam2023gpt, dubey2024llama} has revolutionized the field with their impressive general abilities in semantic understanding, coding, planning, and reasoning.
The strong generalization capabilities has inspired two main approaches in language-conditioned manipulation: pre-training vision-language-action models using large-scale robotics data, as demonstrated by RT-2 and Palm-E~\citep{rt22023arxiv, driess2023palm, padalkar2023open}, and integrating foundation models into agent workflows, like VoxPoser and Copa~\citep{huang2023voxposer, huang2024copa}, which combine LLM/VLM outputs with grasp prediction~\citep{fang2020graspnet, fang2023robust} and motion planning algorithms~\citep{karaman2011sampling}.



While real-world robotics experiments provide valuable insights, their complexity and environmental variability often challenge reproducibility. Simulation-based evaluation has emerged as a fair and practical alternative. Existing benchmarks like RLBench, Calvin, and LIBERO \citep{james2020rlbench, mees2022calvin, liu2024libero} offer diverse task sets but fall short in addressing the unique requirements of foundation model-based methods. Tasks designed to align with the capabilities of foundation model-based algorithms should encompass nuanced semantic understanding of user intent, the integration of common-sense knowledge, and a robust ability to interpret diverse visual scenes, as well as require sophisticated multi-step reasoning. Such tasks demand a sophisticated integration of multimodal understanding to effectively interpret and respond to complex, real-world contexts. For example, one of the tasks in RT-2 \citep{rt22023arxiv} is ``move the Coke can to Taylor Swift", while another task in CoPA \citep{huang2024copa} is ``make me a cup of pour-over coffee". The first task challenges the robot to use common sense to identify Taylor, a capability of knowledge transfer that previous policies struggled to achieve. The second task further intensifies the difficulty, requiring the robot to decompose the task into subtasks and execute the steps to operate a coffee machine—a long--horizon challenge that has previously been difficult for a single policy to accomplish.

To better define the types of language-conditioned manipulation tasks suited for foundation models and provide a standardized evaluation suite to advance robotics research, we introduce VLABench. VLABench is an open-source benchmark specifically designed for methods utilizing foundation models. The tasks in VLABench are carefully divided into several dimensions to evaluate models across various aspects, including 1) Mastery of \textbf{common sense and world knowledge}, 2) Understanding of \textbf{mesh and texture}, 3) Comprehension of \textbf{semantically rich instructions}, 4) \textbf{Spatial} understanding, 5) Grasp of \textbf{physical rules}, and 6) \textbf{Reasoning} ability. For benchmarking purposes, VLABench offers 100 task categories with comprehensive evaluations across various methods. With a diverse collection of over 2,000 3D objects and scenes, VLABench creates a wide range of visual contexts and tasks. It enables the assessment of generalization capabilities through learning across multiple skills, providing thorough evaluations spanning visual, linguistic, planning, knowledge transfer, and action dimensions.

To ensure fair comparison and evaluation, we develop an automated data collection framework to construct standardized datasets for each task, supporting model training and fine-tuning. Using this dataset, we conduct extensive experiments to evaluate and benchmark three distinct types of approaches: pre-trained VLA, workflows integrating foundation models, and vision-language models (VLMs). The experimental results indicate that existing VLA methods perform poorly on our tasks and don’t exhibit the level of generalization abilities or the “emergent” phenomena observed in large models \citep{wei2022emergent}. We summarize contributions as follows:

\begin{itemize} 
\item We propose VLABench, the first benchmark designed to comprehensively evaluate the capabilities of VLAs and VLMs in robotics manipulation tasks, covering multiple dimensions such as skills, vision, language, task execution, common sense, and reasoning.

\item We define 100 novel LCM tasks tailored to the capabilities of foundation models within a standardized evaluation framework. These tasks require a deep understanding of semantics, vision, spatial reasoning, and physical laws, as well as the ability to plan long-horizon tasks and transfer world knowledge and common sense into task execution.

\item We provide a scalable data construction framework and a standardized evaluation dataset. This automated data construction approach facilitates future research on pretraining robotics data.

\item Our experiments demonstrate that current pre-trained VLAs have yet to exhibit the strong generalization capabilities observed in LLMs, and existing SOTA VLMs also show limitations in embodied scenarios.
 \end{itemize}

\section{Related Works}

\begin{figure*}
    \centering
    \includegraphics[width=1\linewidth]{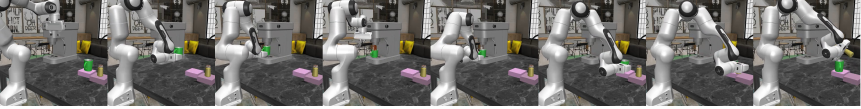}
    \caption{\textbf{Long horizon task requiring reasoning.} This task involves a request for a latte in an interactive scenario. The agent needs to recognize the requirement for coffee with milk and integrate multiple skills, including picking, placing, tool use, pressing, and pouring.}
    \label{fig:long-horizon}
    \vspace{-1em}
\end{figure*}

\begin{table*}[htbp]
  \centering
  \renewcommand{\arraystretch}{1.0}  
    \resizebox{\textwidth}{!}{
    \begin{tabular}{>{\raggedright\arraybackslash}ccccccccccccc}
    \toprule
    Benchmarks & SemLang & LogiReason & Knowledge & DR & N-task & Cate-obj & N-obj & AI-Gen & MultiCam & PCD & Cross Emb & Auto Traj\\
    \midrule
    \textbf{Alfred}\citep{shridhar2020alfred} & {\color{red}\ding{55}} & {\color{red}\ding{55}} & {\color{red}\ding{55}} & {\color{red}\ding{55}}& 7 & - & 3578 & {\color{red}\ding{55}} & {\color{red}\ding{55}} & {\color{red}\ding{55}} &  {\color{green}\checkmark} & {\color{red}\ding{55}}\\
    \textbf{Rlbench}\citep{james2020rlbench} & {\color{red}\ding{55}} & {\color{red}\ding{55}} & {\color{red}\ding{55}} & {\color{red}\ding{55}} &100 & 28 & 28 & {\color{red}\ding{55}} & {\color{green}\checkmark} & {\color{red}\ding{55}} & {\color{red}\ding{55}} & {\color{green}\checkmark}\\
    \textbf{Calvin}\citep{mees2022calvin} & {\color{red}\ding{55}} & {\color{red}\ding{55}} & {\color{red}\ding{55}} & {\color{red}\ding{55}} & 34 & 5 & 30 & {\color{red}\ding{55}} & {\color{green}\checkmark} & {\color{red}\ding{55}} & {\color{red}\ding{55}} & {\color{red}\ding{55}}\\
    \textbf{ManiSkill}\citep{gu2023maniskill2, mu2021maniskill, taomaniskill3} & {\color{red}\ding{55}} & {\color{red}\ding{55}} & {\color{red}\ding{55}} & {\color{green}\checkmark} & 20 & 100 & 2600 & {\color{red}\ding{55}} & {\color{green}\checkmark} & {\color{green}\checkmark} & {\color{green}\checkmark} & {\color{green}\checkmark} \\
    \textbf{LIBERO}\citep{liu2024libero} & {\color{red}\ding{55}} & {\color{red}\ding{55}} & {\color{red}\ding{55}} & - & 130 & 51 & 75 & {\color{red}\ding{55}} & {\color{red}\ding{55}} & {\color{red}\ding{55}} & {\color{red}\ding{55}} & {\color{red}\ding{55}}\\
    \textbf{RoboCASA}\citep{robocasa2024} & {\color{red}\ding{55}} & {\color{red}\ding{55}} & {\color{red}\ding{55}} & {\color{green}\checkmark} & 100 & 153 & 2509 & {\color{green}\checkmark} & {\color{green}\checkmark} & {\color{red}\ding{55}} & {\color{green}\checkmark} & {\color{green}\checkmark}\\
    \textbf{ARNOLD}\citep{gong2023arnold} & {\color{red}\ding{55}} & {\color{red}\ding{55}} & {\color{red}\ding{55}} & -  & 8 & - & 40 & {\color{green}\checkmark} & {\color{green}\checkmark} & {\color{green}\checkmark} & {\color{red}\ding{55}} & {\color{green}\checkmark}\\
    \textbf{Behavior-1K}\citep{li2023behavior} & {\color{red}\ding{55}} & {\color{red}\ding{55}} & {\color{red}\ding{55}} & {\color{green}\checkmark} & 1000 & 2211 & 9331 & {\color{red}\ding{55}} & {\color{red}\ding{55}} & {\color{red}\ding{55}} & {\color{green}\checkmark} & {\color{red}\ding{55}}\\
    \textbf{Habitat 2.0}\citep{szot2021habitat} & {\color{red}\ding{55}} & {\color{red}\ding{55}} & {\color{red}\ding{55}} & - & 3 & 46 & 169 & {\color{red}\ding{55}} & {\color{red}\ding{55}} & {\color{red}\ding{55}} & {\color{red}\ding{55}} & {\color{red}\ding{55}}\\
    \midrule
    \textbf{VLABench} & {\color{green}\checkmark} & {\color{green}\checkmark} & {\color{green}\checkmark} & {\color{green}\checkmark} & 100 & 163 & 2164 & {\color{green}\checkmark} & {\color{green}\checkmark} & {\color{green}\checkmark} & {\color{green}\checkmark} & {\color{green}\checkmark} \\
    \bottomrule
    \end{tabular}
    }
    \vspace{-.5em}
    \caption{Comparison of Popular Benchmarks in Robot Learning. \textbf{SemLang}: Semantically rich language instructions. \textbf{LogiReason}: Task logic and relevant information reasoning. \textbf{Knowledge}: Tasks require the application of common sense and world knowledge. \textbf{DR}: Strong task domain randomization. \textbf{N-task}: Total number of tasks. \textbf{Cate-obj}: Categories of assets used in the simulation. \textbf{N-obj}: Total number of objects in the asset library. \textbf{AI-Gen}: Use of generative AI models for asset library creation. \textbf{MultiCam}: Use of multiple cameras. \textbf{PCD}: Support point cloud data in 3D methods. \textbf{Cross Emb}: Support for cross-embodiment. \textbf{Auto Traj}: Supporting automated data collection}
    \label{tab:compare_table}
    \vspace{-1em}
\end{table*}
\noindent\textbf{Benchmarks and Datasets.}
Numerous benchmarks such as RLBench and LIBERO \citep{james2020rlbench, mees2022calvin, liu2024libero, zheng2024robocas, li24simpler} have been proposed to evaluate language-conditioned manipulation policies in realistic physical settings. A comparison of these benchmarks is provided in Table \ref{tab:compare_table}. However, most of these focus on skill learning and fail to sufficiently address long-horizon planning capabilities.
Meanwhile, some benchmarks \citep{shridhar2020alfred, szot2021habitat, yenamandra2023homerobot} address room-scale mobile manipulation tasks that require long-term memory or reasoning. Yet, these interactions typically occur through interfaces, rather than direct physical manipulation, limiting the transferability of learned policies to real-world scenarios. Additionally, while efforts \citep{robocasa2024, li2023behavior, liu2024libero, gu2023maniskill2} have made strides in task format, difficulty, and scale, these benchmarks have largely overlooked the guiding role of language in tasks, often relying on template instructions that explicitly specify the robot’s actions. VLABench is the first to introduce features such as natural human interaction, implicit goal-oriented semantics, and requirements based on common sense into robot manipulation tasks, as shown in Figure \ref{fig:long-horizon}. In terms of generalization evaluation, previous works \citep{james2020rlbench, mees2022calvin, liu2024libero} typically assess models at the instance level within the same category, which limits their ability to evaluate generalization across diverse object categories or different tasks within the same skill set. In contrast, VLABench is the first benchmark to evaluate generalization capabilities across a wide range of tasks, object types, and task categories, providing a more comprehensive assessment of model versatility.

Large-scale datasets have been built in both real and simulation \cite{padalkar2023open, walke2023bridgedata, robocasa2024, brown2020language} for large-scale imitation learning for manipulation. However, real-world data faces challenges related to scalability, making it difficult to gather sufficient data at scale \citep{brohan2022rt}. Simulated datasets, while more scalable, often suffer from limited diversity in scenarios and tasks \cite{james2020rlbench, mees2022calvin}, and still require teleoperation \cite{liu2024libero, gong2023arnold} for data collection. VLABench addresses these limitations by offering a broader range of tasks that are more closely aligned with real-world conditions, covering diverse aspects of vision, language, tasks, and skills. Furthermore, it introduces an efficient and robust process for the automated generation of simulated data, significantly enhancing task diversity and scalability.

\noindent\textbf{Pretrained Vision-Language-Action Models.}
The recent rise of multimodal models \citep{achiam2023gpt, zeng2023clip2, dosovitskiy2020image, liu2023llava} and the collection and organization of operational datasets\cite{o2023open, walke2023bridgedata}, have led to the integration of vision-language-action models (VLAs) \citep{brohan2022rt, driess2023palm, rt22023arxiv, kim2024openvla} into language-conditioned manipulation tasks. While the term VLA generally refers to models that combine visual and language inputs for policy learning, we focus specifically on approaches leveraging pre-trained models. Several works \citep{driess2023palm, brohan2022rt, kim2024openvla} have applied further training to pre-trained vision-language models (VLMs) for language-conditioned manipulation. These models demonstrate impressive generalization to unseen objects and tasks, yet their control precision is somewhat limited by the discretization of actions \citep{pearce2023imitating}. To address this limitation, some approaches have explored using diffusion models \citep{pearce2023imitating, chi2023diffusion} as policy networks or using diffusion decoders \citep{li2024cogact, wen2024tinyvla}. Pre-trained models based on diffusion models \citep{liu2024rdt, lin2024data} have shown promising advancements in improving continuous space distribution learning. VLABench includes a selection of these representative methods for comprehensive evaluation.

\noindent\textbf{Framework Utilizing Foundation Models.}
Pre-trained language models\cite{brown2020language, Sun2024MOSS, dubey2024llama} and vision-language models\citep{achiam2023gpt, liu2024visual} have demonstrated strong generalization and versatility. Some researchers\cite{huang2023voxposer, huang2024copa, liu2024moka} combine the general perception and cognitive abilities of these pre-trained models with traditional planning and control algorithms to create agent workflows.
These frameworks allow robots to perform complex zero-shot manipulation tasks without requiring additional training.
To harness the capabilities of foundation models for manipulation, some works \cite{liang2023code, huang2023voxposer} utilize the code comprehension and generation abilities of large language models alongside motion planning optimization algorithms to tackle fundamental manipulation tasks.
Additionally, some methods\cite{huang2024copa, hu2023look} leverage large models to decompose long-horizon tasks into subtasks, then integrate perception and trajectory generation modules to construct the entire manipulation pipeline. However, most zero-shot methods of this kind heavily rely on prompt design \cite{huang2023voxposer}, the accuracy of each module, and even the specific parameters of the models being invoked \citep{huang2024copa}. While these methods demonstrate strong generalization, they often face challenges with accuracy. VLABench provides a zero-shot evaluation framework to assess the performance of such workflows and offers insights into their effectiveness.

\section{VLABench}

\subsection{Task Description}
\label{sec:task_des}
\begin{figure*}
    \centering
    \includegraphics[width=1\linewidth]{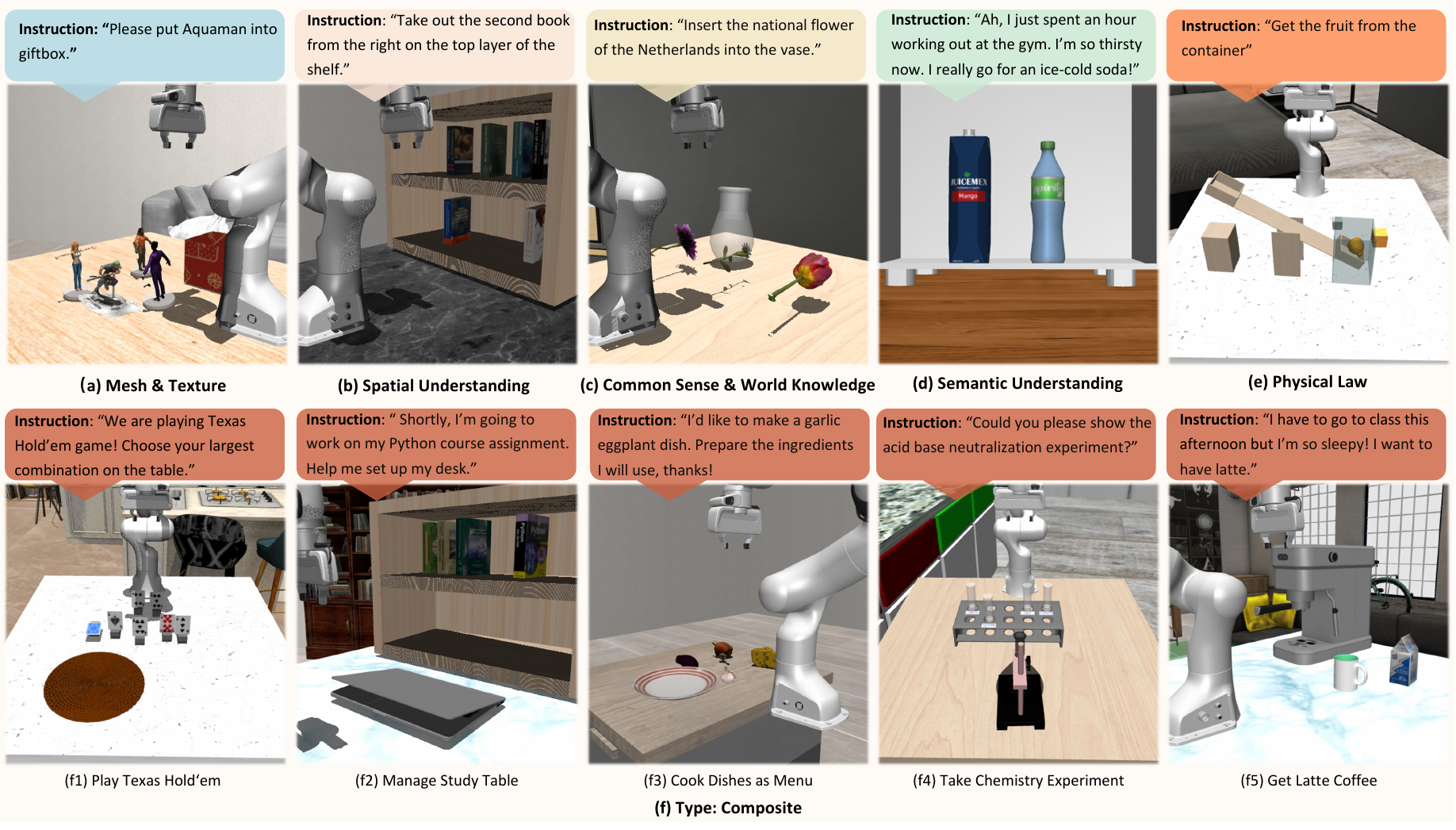}
    \caption{\textbf{Task examples in each dimension.} The first row showcases examples of primitive tasks from Section \ref{sec:task_des}, while the second row presents examples of composite tasks.}
    \label{fig:task_example}
    \vspace{-1em}
\end{figure*}

VLABench is composed of 60 primitive and 40 composite tasks, categorized by task difficulty and required timesteps. These tasks are designed to encompass a rich variety of skills while covering ample visual and language semantic information. For skill learning, 100 tasks in VLABench cover a wide range including 1) Pick\&place, 2) Open\&close door, 3) Open\&close drawer, 4) Hang objects on the wall, 5) Use tool e.g. Hammer nail, 6) Press button, 7) Insert, 8) Pour, 9) Twist, and 10) Explore. In addition, VLABench places greater emphasis on real-life scenarios and essential daily tasks, representing more interactive language instructions, a wider variety of task settings, the integration of common sense and societal knowledge, and long-horizon tasks requiring logical planning, as shown in Figure \ref{fig:task_example}. Notably, VLABench adopts a stricter definition of task generalization, which will be elaborated on in the Section \ref{sec:benchmark}. The whole task list can be found in supplementary material. 

\noindent\textbf{Primitive Tasks.}
Primitive tasks are divided into five dimensions, each corresponding to the assessment of a specific ability dimension.
\begin{itemize}
    \item \textbf{Mesh\&Texture Understanding.} This type of task requires the model to recognize different meshes and understand various texture features. Take the \textit{SelectToy} task shown in Figure \ref{fig:task_example} (a) as an example, the robot is directly required to place a specific toy e.g. Aquaman into a receptacle. The model must possess strong visual capabilities to accurately recognize such complex meshes and textures.
     \item \textbf{Spatial Understanding.} Spatial understanding tasks involve various spatial relationships, such as the nth left/right position, inside/outside of a receptacle, the m\textit{th} row and n\textit{th} column, near/far, and beside a specific object, representing relative positional relationships. Figure \ref{fig:task_example} (b) shows one case in task in \textit{PullBook}. Such a complex relative positional relationship imposes extremely high demands on the model’s multimodal understanding \cite{cheng2024spatialrgpt}.
    \item \textbf{Common Sense \& World Knowledge.} Tasks relative to common sense/world knowledge require the agent to transfer the knowledge gained in the pre-train stage to solve the problem. The task shown in Figure \ref{fig:task_example} (c) requires the agent not only to recognize different types of flowers from visual information but also to leverage world knowledge to determine that ``the tulip is the national flower of the Netherlands".
    \item \textbf{Semantic Understanding.} This type of task emphasizes the complexity, subtlety, and natural interactivity of language instructions. Task objectives are often implicitly conveyed through a natural conversation. To perform well in \textit{GetDrink} task as Figure \ref{fig:task_example} (d) shown, the agent must capture the implied request from a lengthy instruction: to take out a chilled cola from the refrigerator.
    \item \textbf{Physical Law.} This type of task expects the robot to integrate visual information and take correct actions based on physical principles and real-time observation. In \textit{UseSeesaw} task in Figure \ref{fig:task_example} (e), the robot is commanded to grasp an object that can not be achieved directly. The agent must recognize the need to apply the principle of leverage by using sufficient weight to lift the target object on the other end. 
\end{itemize}
\noindent\textbf{Composite Tasks.}
Composite tasks in VLABench involve the combination of multiple skills, long-term task planning, and multi-step logical reasoning from instructions, scenes, and even game rules. Figure \ref{fig:task_example} (f) showcases a variety of challenging complex tasks. Composite tasks have a significantly longer trajectory horizon, with an average episode length exceeding 500 timesteps—considerably more than the average of 120 timesteps for primitive tasks. In Figure \ref{fig:task_example} (f1), The agent must not only correctly identify all poker cards from visual information and use world knowledge of poker rules to select the best hand, but also flip face-down cards to acquire complete information. This type of task, requiring the agent to consciously fulfill prerequisite conditions, has never been modeled before. Composite tasks also require extracting the user’s implicit needs from natural dialogue. The case in Figure \ref{fig:task_example} (f2) needs the agent to place the Python textbook on the table and open the laptop, without direct instruction.
\subsection{Benchmark}
\label{sec:benchmark}

\noindent\textbf{Evaluation.}
VLABench organizes evaluations into three main categories: assessments of pretrained or fine-tuned vision-language-action (VLA) models, heuristic workflows that integrate foundation models with various algorithms, and multi-dimensional evaluations of vision-language models (VLMs). 
\begin{itemize}
    \item \textbf{Generalization Ability of VLAs.} For trained vision-language-action (VLA) models, the evaluation in VLABench includes two settings: seen objects and unseen objects. The seen objects evaluation closely aligns with the data distribution of the training set, primarily testing the model’s skill acquisition. Meanwhile, the unseen objects evaluation presents a greater challenge, requiring the model to exhibit strong generalization capabilities. Unlike previous benchmarks \cite{james2020rlbench, mees2022calvin}, VLABench defines unseen objects as entirely different categories. For instance, in the \textit{PickFruit} task, target objects for seen evaluation include apples, bananas, pears, and oranges, while unseen objects include kiwis, mangos, strawberries, lemons, and other distinct fruits. This setup requires the model to demonstrate not only strong visual generalization capabilities but also to handle the vastly differing common-sense knowledge associated with different categories of objects, as well as the challenge of processing lengthy instructions with unfamiliar tokens.
    \item \textbf{Zero-shot Transfer Ability of Heuristic Workflow.} Training-free workflow methods are evaluated under a single setting but in many ability dimensions. Apart from the capability points for primitive tasks mentioned in Section \ref{sec:task_des}, we extend the evaluation to cover various skills and long-horizon tasks to assess the overall capability and execution robustness of the workflow.
    \item \textbf{Comprehensive Evaluation of VLMs’ Capabilities.} Similar to heuristic workflows, the evaluation of VLMs is also comprehensive. Since VLMs lack intrinsic action capabilities, we organized a skill library and integrated it into a domain-specific language (DSL) \cite{nordmann2014survey, sutherland2019robolang}, leveraging annotated asset information as prior knowledge. This DSL functions as a straightforward API that VLMs can call, enabling efficient interaction. The whole evaluation pipeline will be discussed in Section \ref{sec:eval vlm}.
\end{itemize}

\noindent\textbf{Metric.} 
Our evaluation focuses on generalization capabilities, but the task success metric, limited to a 0/1 score, is better suited for assessing straightforward skill learning. Thus, we introduce Progress Score (PS) as a graduated metric for more nuanced assessment. The computation equation of PS is: 
\begin{equation}
PS = \alpha \cdot \frac{n_{correct}}{N} + (1 - \alpha) \cdot \frac{m_{done}}{M}
\end{equation}
where $N$ indicates the total number of target objects and receptacles, $n_{correct}$ is the number of those selected correctly. $M$ represents the total number of sub-steps in the task, with $m_{done}$ indicating the number of completed sub-steps. Here, $\alpha$ is the weight assigned to correct decisions, default set to 0.2, while $1 - \alpha$ represents the weight assigned to task progress. For the evaluation of VLMs, we employed a more detailed scoring method, the metric includes Skill Recall Rate, Parameter Recall Rate, Skill\&Parameter Recall Rate, and Precise Matching Rate. Please refer to Section \ref{appendix:eval vlm} for further details.

\subsection{Simulation}
\label{sec:simulation}
\noindent\textbf{Simulator.} VLABench is built based on Mujoco\citep{todorov2012mujoco} and its control suite dm\_control\citep{tunyasuvunakool2020}. We selected Mujoco as the core simulation platform for our benchmark due to its lightweight design, high performance, and exceptional physical realism. These advances enable convenient, rapid evaluation of diverse algorithms. The VLABench framework is highly modular, meaning various object entities can be flexibly combined to create large-scale and diverse tasks and scenarios. 

\noindent\textbf{Assets.} 
To meet the requirements of diverse tasks and capability assessments, we built an asset library centered around multiple task themes. We inherited some annotated assets from Robocasa \citep{robocasa2024} and retrieved numerous 3D models from Objaverse\citep{deitke2023objaverse}. For novel tasks, such as the series of tasks we created around the toy theme, we carefully gathered a variety of high-quality character models from online 3D model sites. These models were then converted to MJCF format using the obj2mjcf \citep{obj2mjcf} tool. Similarly to previous work \citep{robocasa2024, li2023behavior}, we expanded the dataset of common simple objects using generative AI models. Specifically, we utilized Tripo.aI’s text-to-3D and image-to-3D features to construct additional 3D objects, and Runaway.ai to generate multiple material textures. 
Ultimately, the asset library we constructed contains 163 categories of objects, totaling 2164 items. Most of the assets are listed in Section \ref{appendix:assets}.

\noindent\textbf{Robots.} 
To ensure versatility and broad applicability, we integrated a range of embodiment types. These include, but are not limited to, various models of 6-axis and 7-axis robotic arms, dual-arm robots, and humanoid robots. In the standard evaluation process, VLABench employs a 7-DoF Franka Emika Panda manipulator equipped with a parallel gripper. We represent the position and orientation of the robot’s end-effector in Euclidean space \( \mathbb{R}^3 \) using 3D coordinates for position and quaternions for orientation. Using inverse kinematics, we then resolve these end-effector poses into the corresponding rotational angles for the seven joints. 

\subsection{Dataset Construction}
\noindent\textbf{Domain Randomization.}
To ensure data diversity and richness, we implemented various types of domain randomization. These randomizations include object position and orientation, mesh scale, scene layout, background and object textures (such as walls, floors, and tabletops), as well as lighting parameters. Details can be found in Section \ref{sec:domain_randomization}.

\noindent\textbf{Trajectory Generation.} 
As human teleoperation is time-consuming and not scalable \citep{liu2024libero, robocasa2024}, we developed an efficient, scalable automated data collection pipeline based on our custom skill library. Inspired by \cite{gong2023arnold}, our data collection framework leverages the prior information including point clouds of the environment, entities' grasp-points, target entity at the current step, etc. The data collection framework includes multiple task-specific motion planners. These motion planners call upon the skills in the skill library based on the current task progress and determine parameters by incorporating prior information. Subsequently, the selected skills generate trajectories using RRT \cite{karaman2011anytime}, with quaternion interpolation achieved through Spherical Linear Interpolation (SLERP). The final trajectory is smoothed using a Bezier curve to optimize path quality. To enhance sample efficiency during data collection, reject sampling and failure-triggered early termination are applied.

\noindent\textbf{Instruction Augmentation.}
We use GPT-4 \cite{achiam2023gpt} to generate descriptions that incorporate target-specific characteristics and interactive instructions that encompass a variety of contexts and intentions. The supplementary material provides details on the generation process and the complete prompts.

\section{Experiments}
Following Section \ref{sec:benchmark}, we conducted experiments centered on pre-trained VLA models, workflows incorporating multiple algorithmic modules, and various VLMs. The remainder of this section provides a detailed description of the experimental setup.
\subsection{Generalization Ability of VLAs}

\begin{table*}[h!]
    \centering
    \renewcommand{\arraystretch}{1.2}    
    \begin{subtable}{\textwidth}
    \centering
    \resizebox{\textwidth}{!}{
    \begin{tabular}{cccccccccccccccccccc}
        \hline
        \textbf{Model} & \multirow{2}{*}{\textbf{Task Name}} & \multicolumn{2}{c}{Add Condiment} & \multicolumn{2}{c}{Insert Flower} & \multicolumn{2}{c}{Select Book} & \multicolumn{2}{c}{Select Drink} & 
        \multicolumn{2}{c}{Select Toy} &
        \multicolumn{2}{c}{Select Tube} &
        \multicolumn{2}{c}{Select Painting} &
        \multicolumn{2}{c}{Select Fruit} &
        \multicolumn{2}{c}{Average} \\ 
        & & \textbf{Seen} & \textbf{Unseen} & \textbf{Seen} & \textbf{Unseen} & \textbf{Seen} & \textbf{Unseen} & \textbf{Seen} & \textbf{Unseen} &
        \textbf{Seen} & \textbf{Unseen} & \textbf{Seen} & \textbf{Unseen} & \textbf{Seen} & \textbf{Unseen} & \textbf{Seen} & \textbf{Unseen} & \textbf{Seen} & \textbf{Unseen} \\ \hline
        \multirow{2}{*}{Octo} & Base & 3.08 & 3.08 & 1.54 & 0.00 & 0.00 & 1.54 & 0.00 & 0.00 & 0.00 & 0.00 & 1.54 & 0.00 & 6.15 & 1.54 & 0.00 & 0.00 & \textbf{1.34} & \textbf{0.77} \\ 
        & Common Sense & 1.54 & 3.08 & 0.00 & 0.00 & 0.00 & 0.00 & 0.00 & 0.00 & 3.08 & 1.54 & 1.54 & 3.08 & 3.08 & 0.00 & 0.00 & 0.00 & \textbf{1.16} & \textbf{0.96}\\
        \hline
        \multirow{2}{*}{OpenVLA} & Base & 12.38 & 8.23 & 13.85 & 7.69 & 7.69 & 4.62 & 8.46 & 4.61 & 3.08 & 4.62 & 7.69 & 6.15 & 40.20 & 28.26 & 4.62 & 3.07 & \textbf{11.74} & \textbf{7.93}\\ 
        & Common Sense & 8.23 & 3.08 & 9.24 & 4.61 & 0.00 & 0.00 & 8.46 & 4.61 & 0.00 & 0.00 & 6.15 & 3.08 & 34.06 & 25.48 & 1.54 & 0.00 & \textbf{8.46} & \textbf{5.11}\\
        \hline
        \multirow{2}{*}{RDT-1B} & Base & 21.54 & 14.46 & 21.54 & 16.92 & 3.08 & 1.54 & 7.69 & 3.08 & 7.69 & 4.62 & 12.38 & 6.15 & 35.16 & 19.72 & 13.85 & 6.15 & \textbf{15.37} & \textbf{9.08}\\ 
        & Common Sense & 16.92 & 4.61 & 14.46 & 3.08 & 0.00 & 0.00 & 7.69 & 0.00 & 4.62 & 1.54 & 7.69 & 0.00 & 32.08 & 16.64 & 12.32 & 3.07 & \textbf{11.97} & \textbf{3.61}\\
        \hline
    \end{tabular}}
    \caption{Evaluation of visual generalization and knowledge transfer.}
    \label{subtable:exp1}
    \vspace{-0.2cm}
    \end{subtable}

    \bigskip
    \begin{subtable}{0.32\textwidth}
    \centering
    \resizebox{\textwidth}{!}{
    \begin{tabular}{p{3cm}ccc}
        \hline
        \textbf{Task} & Octo & OpenVLA & RDT-1B\\ \hline
        Add Condiment & 0.00 & 0.00 & 6.15 \\
        Insert Flower & 0.00 & 10.00 & 9.24 \\
        Select Drink & 0.00 & 7.69 & 3.08 \\
        Select Toy & 0.00 & 0.00 & 3.08 \\
        Select Tube & 0.00 & 3.08 & 0.00 \\
        Average& \textbf{0.00} & \textbf{4.15} & \textbf{4.31} \\
        \hline 
    \end{tabular}
    }
    \caption{Evaluation of language instruction generalization.}
    \label{subtable:exp2}
    \vspace{-0.2cm}
    \end{subtable}
    \hfill
    \begin{subtable}{0.32\textwidth}
    \centering
    \resizebox{\textwidth}{!}{
    \begin{tabular}{p{3cm}ccc}
        \hline
        \textbf{Task} & Octo & OpenVLA & RDT-1B\\ \hline
        Select Poker & 0.00 & 7.69 & 4.62 \\
        Select Majhong & 0.00 & 4.62 & 3.07 \\
        Select Billiards & 0.00 & 3.07 & 4.62 \\
        Select Ingredient & 0.00 & 0.00 & 0.00 \\
        Friction QA & 0.00 & 10.46 & 6.92 \\
        Average& \textbf{0.00} & \textbf{4.46} & \textbf{3.85} \\
        \hline 
    \end{tabular}
    }
    \caption{Evaluation of unseen but similar task generalization.}
    \label{subtable:exp3}
    \vspace{-0.2cm}
    \end{subtable}
    \hfill
    \begin{subtable}{0.31\textwidth}
    \centering
    \resizebox{\textwidth}{!}{
    \begin{tabular}{p{3cm}ccc}
        \hline
        \textbf{Task} & Octo & OpenVLA & RDT-1B\\ \hline
        Find Unseen Object & 0.00 & 7.69 & 0.00 \\
        Play Texas Holdem & 0.00 & 3.54 & 3.08 \\
        Cluster Toy & 0.00 & 0.00 & 5.06 \\
        Hammer and Hang & 0.00 & 0.00 & 0.00 \\
        Get Latte Coffee & 0.00 & 2.08 & 8.56 \\
        Average& \textbf{0.00} & \textbf{2.66} & \textbf{3.34} \\
        \hline 
    \end{tabular}
    }
    \caption{Evaluation of composite tasks.\newline}
    \label{subtable:exp4}
    \vspace{-0.2cm}
    \end{subtable}
    
    \caption{Overall experiment result of generalization ability of fine-tuned VLAs.}
    \label{table:vla-exp}
    \vspace{-1em}
\end{table*}
Pretrained VLAs are expected to possess robust generalization and versatility similar to LLMs. Experiments about are set to address the following research questions:

\textbf{Q1:} Do pre-trained VLAs exhibit stronger general abilities with unseen categories of objects?

\textbf{Q2:} Can pre-trained VLAs transfer their general knowledge and behavioral abilities to similar but unseen tasks? 

\textbf{Q3:} Can pre-trained VLAs understand natural user interactions and implicit goal requirements? 

\textbf{Q4:} Do pre-trained VLAs have the potential to transfer their world knowledge to related tasks? 

\textbf{Q5:} Can existing VLA architectures accurately support the completion of long-horizon tasks? 

\noindent\textbf{Experiment Setup.}
To investigate the questions outlined above, we fine-tuned various pre-trained VLA architectures, including OpenVLA, Octo, and RDT-1B \citep{kim2024openvla, team2024octo, liu2024rdt}, on our high-quality dataset. Our composite tasks demand generalization across language, vision, common sense, and long-horizon reasoning, requiring the integration of multiple skills. To assess generalization ability, we selected primitive tasks as the foundation for evaluation. Within each category of primitive tasks, the \textit{Mesh\&Texture} (base) tasks, \textit{Common sense \& World knowledge} tasks, and \textit{Semantic} tasks share similar task setups and trajectories. Therefore, we opt for joint training on base and common sense data across each task category and evaluate in different settings. During the fine-tuning stage, we sample 100 trajectories from each task category, resulting in a total of 1,600 trajectories to ensure balanced representation across tasks. For complex tasks, we perform fine-tuning separately within the domain of each task and conduct evaluations independently.

\noindent\textbf{Result and Analysis.}
In the evaluation stage, different task settings are applied to cover multiple generalization abilities. In Table \ref{subtable:exp1}, we present experimental results comparing the generalization capabilities of vision and common sense by evaluating seen and unseen categories of objects. The experimental results indicate that the current large-scale pre-trained VLAs did not exhibit the expected rapid adaptation to downstream tasks. The fine-tuned models performed poorly in primitive tasks especially involving the Pick\&Place skill, the findings are similar to \cite{robocasa2024}. Limited by its discretization process and single-frame input architecture, OpenVLA’s skill-learning capability is lower than that of RDT-1B. However, benefiting from pre-trained VLMs, OpenVLA achieves higher scores than RDT-1B on common-sense tasks involving unseen objects. Our analysis suggests that although OpenVLA only fits trajectory data during pre-training, its foundation on Llama2-7B provides it with greater generalization potential. 

In Table \ref{subtable:exp2}, \ref{subtable:exp3}, and \ref{subtable:exp4}, evaluations were conducted on out-of-domain semantically rich language, unseen but similar tasks, and composite tasks respectively. These experimental results indicate that current architectures and pre-training approaches are insufficient for equipping VLA models with stronger semantic understanding, skill transfer, and long-horizon planning capabilities. Analogous to the classic paradigm of pretraining-finetuning in large language models during the GPT-3 era \citep{qiu2020pre}, it is still difficult to determine how much gain VLA has achieved from pretraining on the scarce, quality-varying dataset of only a few million samples. Moreover, this becomes even more challenging if the backbone has already undergone large-scale vision-language training. Drawing an analogy to the development trajectory of large language models, the present state of VLAs is still far from reaching a level comparable to GPT-2. Further ablation studies and analysis are presented in Section \ref{appendix:ablation vla}. 
\begin{figure*}[t]
    \centering
    \begin{minipage}[t]{0.48\textwidth}
        \centering
        \includegraphics[width=\linewidth]{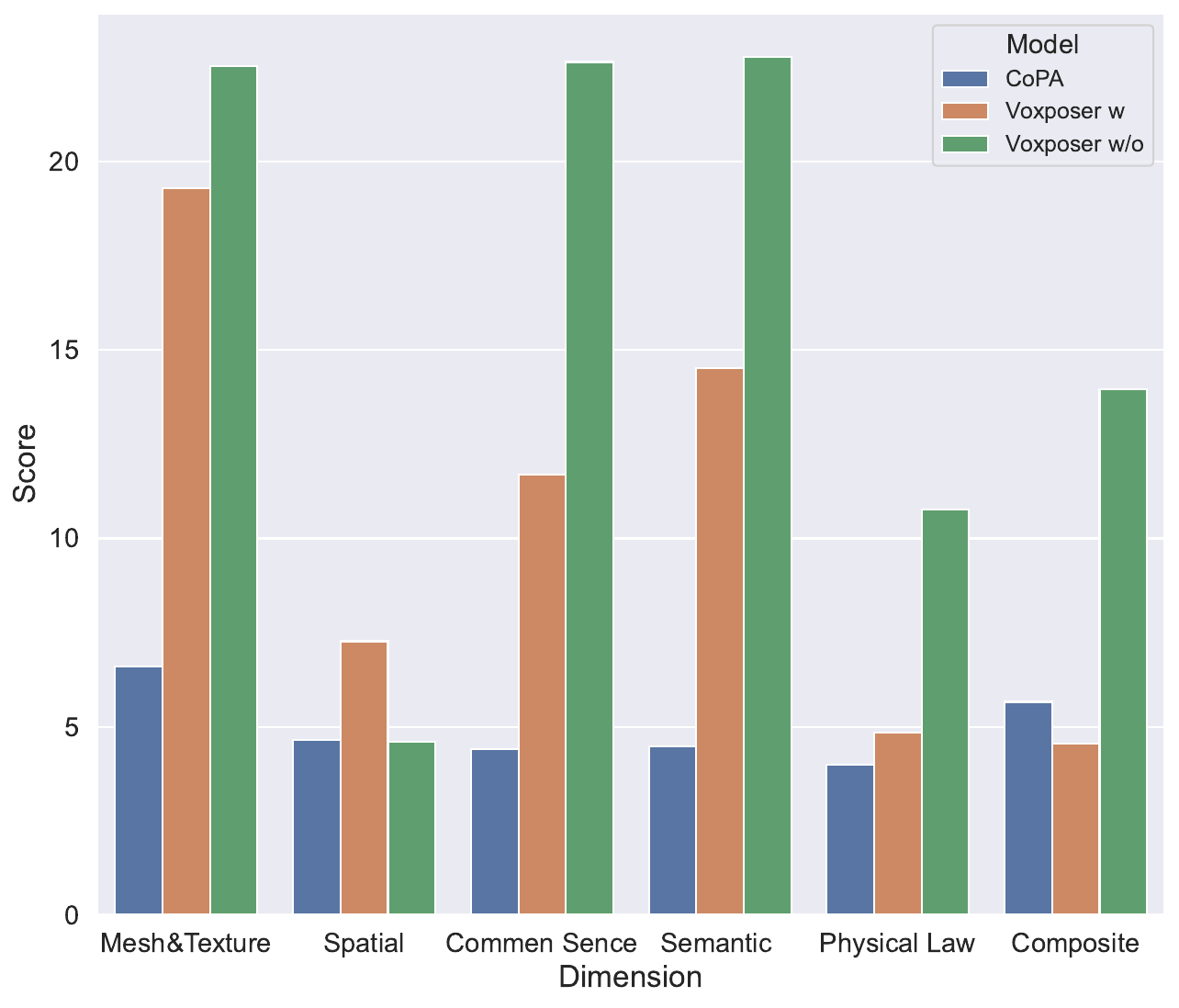}
        \caption{Evaluation results for Voxposer and CoPA.  \textit{Voxposer w/o} refers to the version without visual perception, where ground truth labels are directly provided for object selection. \textit{Voxposer w} uses GPT-4V as the visual perception module.}
        \label{fig:voxposer and copa}
        \vspace{-1em}
    \end{minipage}
    \hfill
    \begin{minipage}[t]{0.48\textwidth}
        \centering
        \includegraphics[width=\linewidth]{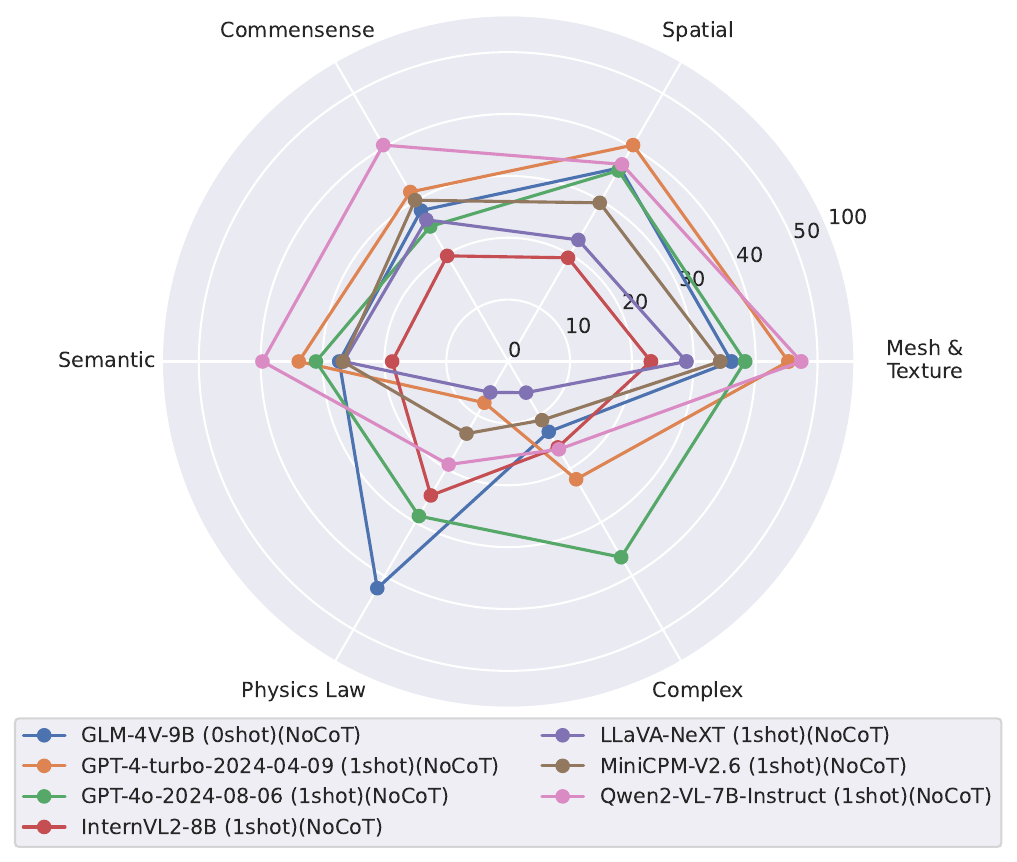}
        \caption{Radar charts depicting the performance of all VLM models across six dimensions. The reason why only GLM-4V-9B is evaluated in a zero-shot setting is that it does not support multi-graph inference, which is required for the other models.}
        \label{fig:all model result}
        \vspace{-1em}
    \end{minipage}
\end{figure*}

\subsection{Performance of Workflow Utilizing Foundation Model}
For our evaluation of foundation model-based algorithms, we reviewed two state-of-the-art frameworks, Voxposer~\citep{huang2023voxposer} and CoPA~\cite{huang2024copa}, and the comparison results are shown in Figure \ref{fig:voxposer and copa}. Given Voxposer's dependence on large language models (LLMs), we assessed Voxposer's performance with and without visual perception capabilities. While Voxposer performed adequately on basic tasks and achieved the Progress Scores of 30–40, its reliance on LLM-driven motion planning often led to grasping failures due to limited information for effective grasp planning, especially when interpreting rotation in non-visual contexts, resulting in low overall scores.

Interestingly, the foundational LLM alone maintained relatively stable scores in semantic understanding and reasoning tasks without visual input. However, adding visual perception slightly reduced performance in these areas while significantly improving spatial reasoning, where LLM-only setups struggle with spatial accuracy due to lack of spatial information.

The lack of closed-loop feedback limits these models' ability to perform physical reasoning tasks, particularly those involving dynamic interactions, leading to lower scores in this dimension. Both models struggle with high-complexity tasks, succeeding mainly in entity recognition but performing poorly in long horizon task reasonable breaking down. This finding underscores the requirements for advancing foundational model-based frameworks to address complex reasoning. In fact, although the methods mentioned above emphasize their zero-shot capabilities and generalization to new scenarios, their modular design often limits the upper bound of their performance. A more detailed discussion of this can be found in Section \ref{appendix:ablation agent}.

\begin{figure*}
    \centering
    \includegraphics[width=1\linewidth]{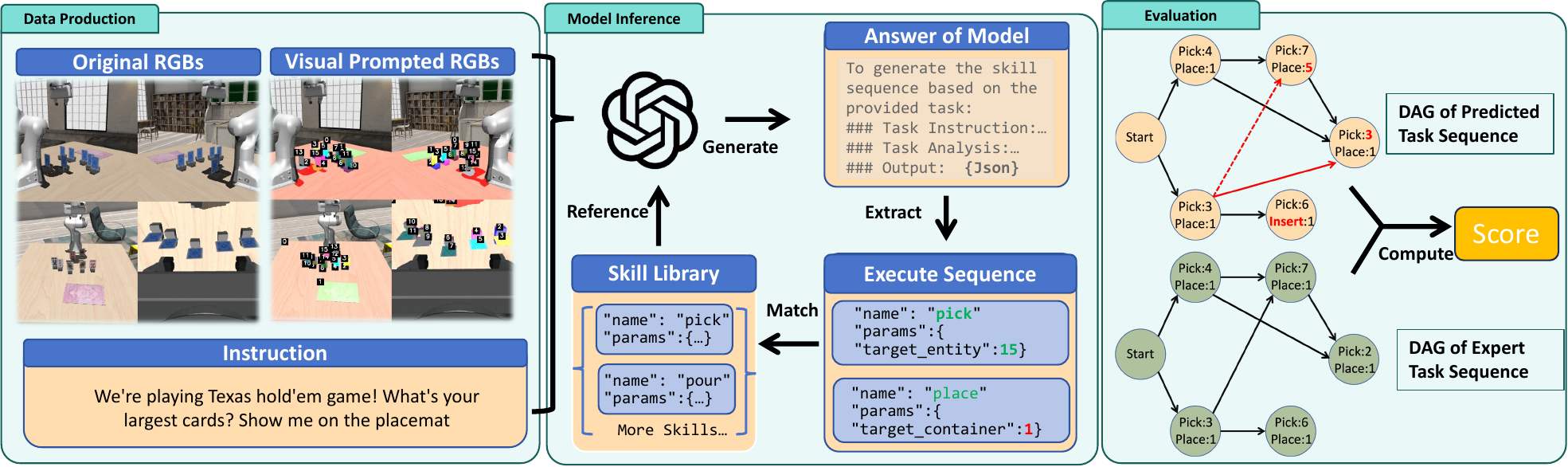}
    \caption{\textbf{Evaluation pipeline for VLMs}. \textbf{Step 1:} Sample the required four-view images, as well as those segmented with numerical information, from the simulation. Meanwhile, save the corresponding instructions and operation sequences. \textbf{Step 2:} Input the original and annotated images and the instructions into the model. Then, obtain the model’s output and extract the generated operation sequence. \textbf{Step 3:} The two operation sequences are evaluated using four metrics, which are weighted and summed to produce the final score.The \textcolor{red}{red} part represents the error output of the model, where the red solid arrows represent the dependencies generated by the error, and red dashed arrows represent the dependencies lost by the error.}
    \label{fig: VLM evaluation pipeline main}
    \vspace{-1em}
\end{figure*}

\subsection{Comprehensive  Ability of VLMs}
\label{sec:eval vlm}

We referred to the evaluation results of multiple series of Vision-Language Models (VLMs) provided by OpenCompass \citep{2023opencompass} and selected several models from different families with strong overall performance. These models include: GPT-4-turbo-2024-0409, GPT-4o-2024-08-06 \cite{achiam2023gpt}, GLM-4V-9B \cite{glm2024chatglm}, MiniCPM-V2.6 \citep{hu2024minicpm}, Qwen2-VL-7B \cite{bai2023qwen}, InterVL2-8B \citep{chen2024internvl}, and LLaVA-NeXT \citep{liu2024llavanext}. We evaluate the comprehensive performance of these models with the dataset derived from naturally self-contained information within a simulated environment. This dataset consists of a complex set of tasks designed to assess the VLM’s ability to perceive visual stimuli and comprehend verbal instructions. There are two types of evaluation approaches for VLMs: interactive and non-interactive. In the following, we will provide a detailed introduction to both approaches.

\noindent\textbf{Non-interactive Evaluation.}
Figure \ref{fig: VLM evaluation pipeline main} illustrates the simplified evaluation process specifically designed for VLMs in VLABench. Firstly the evaluation dataset is generated by initializing a series of task scenarios, each associated with two four-view diagrams: one annotated with masks and labels to identify distinct entity segments, and the other serving as a reference image without annotations, as shown in Data Production module in Figure \ref{fig: VLM evaluation pipeline main}. A randomly selected linguistic instruction from GPT4 relevant to the task accompanies these diagrams, forming the input to the Vision-Language Model (VLM). 

During inference time, we provide a detailed description of the skill library, the requirements of output format, and several few-shot examples in different settings. These elements collectively form the system prompt for querying the VLM. The VLM is required to generate DSL output consisting of a sequence of skills, where each skill includes a name and associated parameters, conforming to predefined patterns to enable systematic evaluation. 

Then, the generated skill sequences are constructed into a directed graph based on their logical dependencies. Subsequently, these DAGs are matched with the reference ones and scored in four metrics. Finally the scores are combined using weighted aggregation to calculate a total score for each model. Please refer to Section 
\ref{appendix:eval vlm} in the supplementary material for more detailed metric computation.

\noindent\textbf{Interactive Evaluation.}
Similar to the VLA and workflow evaluation process mentioned in previous sections, interactive evaluation computes a task progress score based on the interaction with the environment. VLABench provides a controller that parses the DSL action sequences output by the VLM into executable actions, which are then applied in a simulation environment to interact with real-world objects. This approach is one of the key metrics for evaluating robotic manipulation tasks. However, it is more time-consuming compared to non-interactive approaches, and its evaluation dimension is relatively limited, as it cannot distinguish between errors in skill selection and those in parameter generation. 

\noindent\textbf{Performance Comparison and Analysis.}
In line with the evaluation framework outlined in the previous section, we evaluate the models across six dimensions of task performance. The results for each VLM model under the 1-shot\footnote{As GLM-4V-9B does not support multiple image inputs, the 0-shot method is employed.} setting are summarized in Figure \ref{fig:all model result}.

Although these VLMs perform well on most multimodal tasks and even some embodied tasks \citep{2023opencompass, li2024embodied}, their performance, including that of GPT-4o, falls short when faced with more complex scenarios, instructions, and more challenging tasks. We surprisingly find that the open-source model Qwen2-VL-7B-Instruct performed competitively, surpassing GPT-4-turbo-2024-04-09 on certain dimensions. However, all models struggled with complex tasks, especially those requiring long-term task decompose and logic reasoning. Only GPT-4o achieved a score in the reasoning dimension comparable to those in other dimensions, while the other models scored around 20 points. Besides, performance declines significantly when linguistic instructions transition from direct semantics to abstract meanings, as shown in semantic dimension. Different models appear to have distinct areas of expertise, e.g. LLaVA-NeXT exhibits weaker spatial perception ability, GLM-4V-9B excels in spatial and even physical law dimensions but lag in semantic comprehension. More ablations and discussions are in Section \ref{appendix:ablation vlm}. Overall, while the models demonstrate promising capabilities, their understanding and planning in embodied environments remain limited, highlighting the need for further advancements.




\section{Conclusion}
We propose VLABench, a large-scale benchmark designed for tasks with long-horizon and multi-dimensional reasoning. Such reasoning and evaluation involves many dimensions, including from vision to knowledge gained in pretrain stage, implicit semantic goal extracting ability, combining the requirement and interactive scene to make reasonable decision, logical reasoning and the ability to make long horizon plan. One of the most important thing we do is to provide a positive definition of the capabilities that intelligent agents with true cognitive abilities should possess and the tasks they should be able to perform, through the provision of 100 standardized task settings. Additionally, VLABench constructed a scalable automated data collection framework for future's potential larger scale pertaining and a standardized dataset for fair comparison of VLAs, both in the present and in future developments. Our diverse and multiple experiments revealed that current VLAs and VLMs face significant challenges in our tasks, and there remain substantial uncertainties in research on robotics scaling. We hope that VLABench will inspire both the future research on robotics pertaining recipe and promote more robust VLA architectures development.
{
    \small
    \bibliographystyle{ieeenat_fullname}
    \bibliography{main}
}
\clearpage
\setcounter{page}{1}
\maketitlesupplementary
\begin{figure}[t]
\centering
    \includegraphics[width=\linewidth]{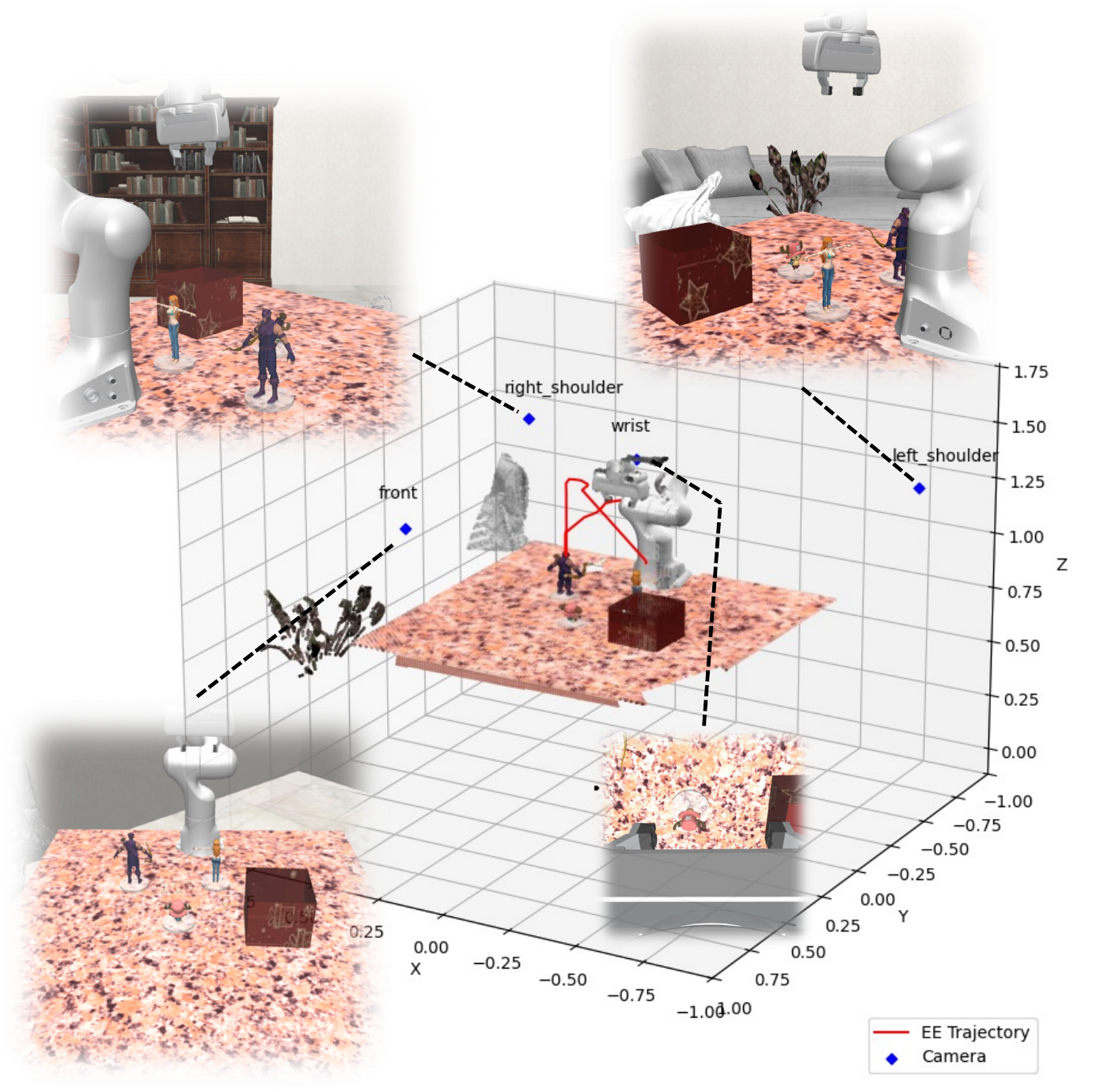}
    \caption{\textbf{Task observations.} The figure illustrates an example data instance from the \textit{Select Toy} task, including multi-camera positions, multi-view RGB images, 3D point clouds, and the expert trajectory.}
    \label{fig:task_obs}
    \vspace{-1em}
\end{figure}

\begin{figure}[t]
    \centering
    \includegraphics[width=\linewidth]{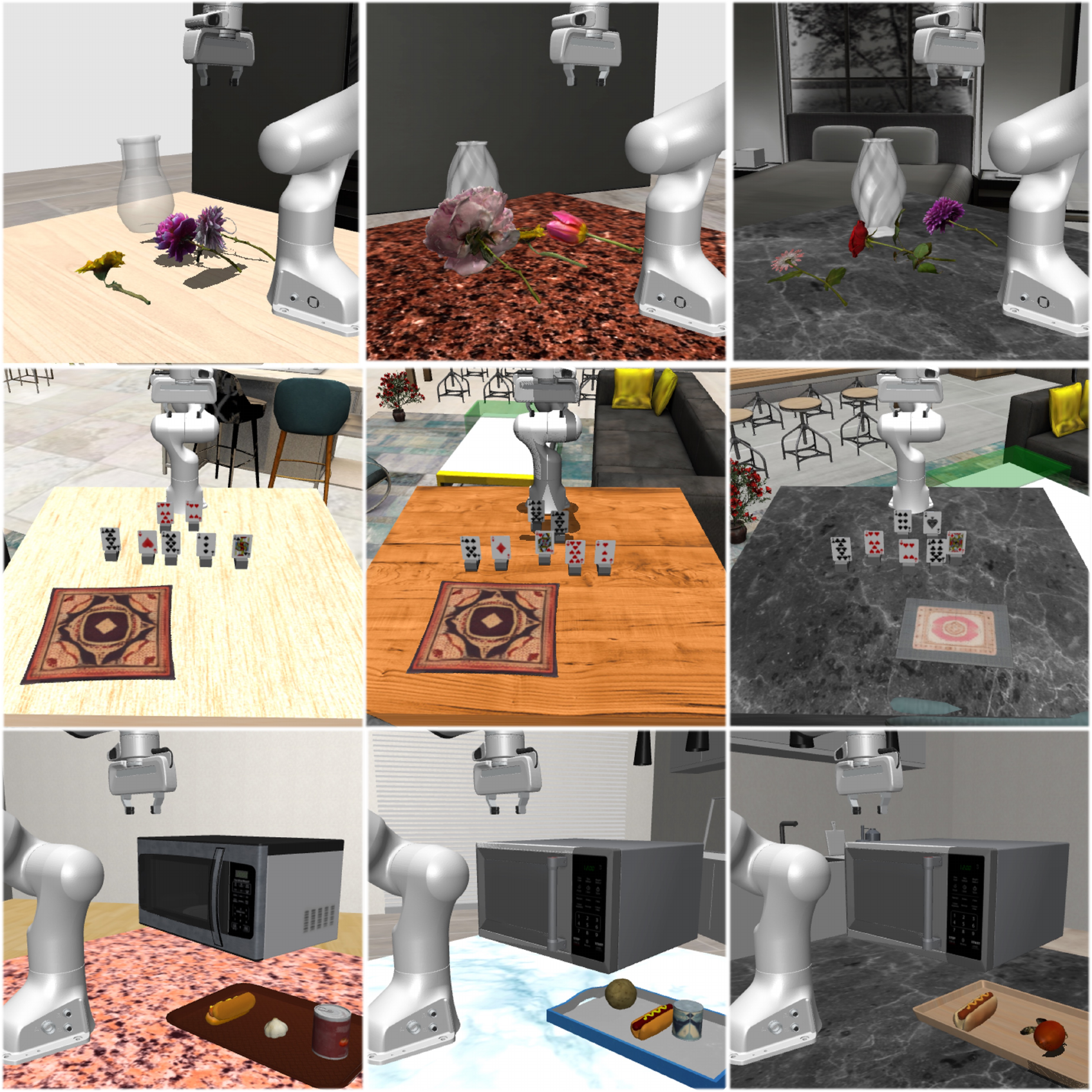}
    \caption{\textbf{Multi cases in the same task.} \textbf{Row 1}: \textit{Insert Flower} task from the left shoulder view. \textbf{Row 2}: \textit{Play Texas Hold’em} task from the front view. \textbf{Row 3}: \textit{Heat Food with Microwave} task from the right shoulder view. Examples in the same row originate from the same task but differ in task objectives, distracting objects, spatial configurations, spatial poses, etc.}
    \label{fig:one-to-many}
    \vspace{-1em}
\end{figure}

\begin{figure}[t]
    \centering
    \includegraphics[width=\linewidth]{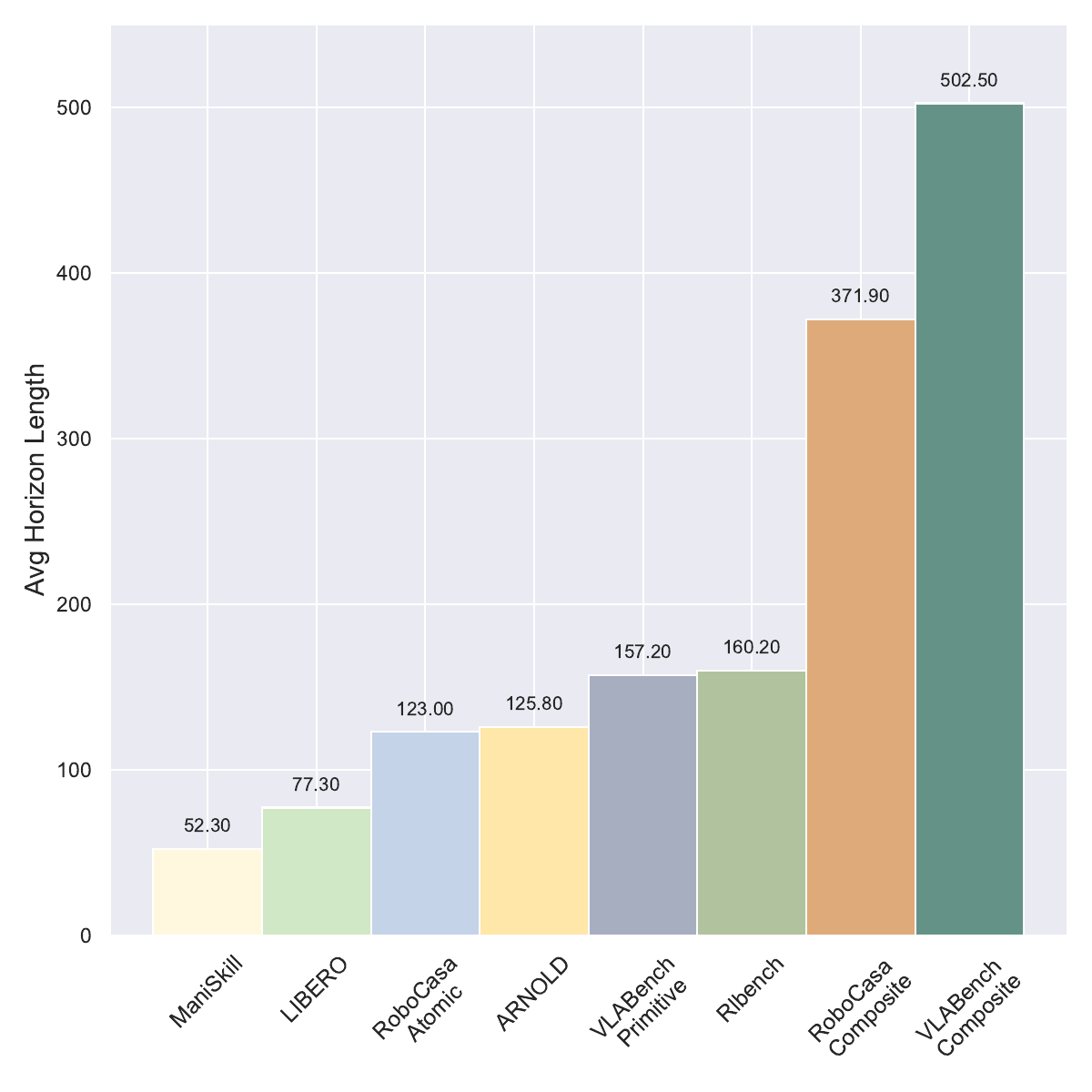}
    \caption{\textbf{Comparison of horizon length.} VLABench significantly surpasses other benchmarks in terms of average task length and reasoning steps.}
    \label{fig:horizon_reason}
    \vspace{-1em}
\end{figure}


\section{Benchmark Implementation}
\subsection{Task Descriptions}
\noindent\textbf{All Tasks.} VLABench includes both 60 primitive and 40 composite tasks. These tasks encompass a wide range of manipulation skills and involve many high-level capabilities. The skills include 1) Pick\&place, 2) Open\&close door, 3) Open\&close drawer, 4) Hang objects on the wall, 5) Use tool e.g. Hammer nail, 6) Press button, 7) Insert, 8) Pour, 9) Twist, and 10) Explore. For higher-level intelligence, VLAbench's evaluation dimensions encompass complex scene understanding, implicit semantic analysis, world knowledge transfer, understanding of physical laws, relative spatial perception, long-term task planning, and even multi-step logical reasoning. Table \ref{table:task_description} provides a detailed introduction to the 100 tasks involved in VLAbench, including the type of each task, the manipulation skills involved, the scope of high-level intelligence examined, the average episode length at a control frequency of 10Hz, as well as a detailed description of the task and an explanation of its challenges. For the sake of clarity in the table, we will use abbreviations to represent the various intelligence dimensions. \textbf{M\&T} corresponds to \textit{Mesh \& Texture Understanding}, \textbf{SP} corresponds to \textit{Spatial Understanding}, \textbf{C\&W} corresponds to \textit{Common Sense \& World Knowledge}, \textbf{SEM} corresponds to \textit{Semantic Conversation Understanding}, \textbf{PHY} corresponds to \textit{Physical Laws Understanding}, and \textbf{L\&R} corresponds to \textit{(Logistic) Reasoning}.

\noindent\textbf{Long-horizon Design with Multistep Reasoning.} Compared to previous benchmarks, VLAbench places more emphasis on comprehensive long-term reasoning. The reasoning defined here includes associating world knowledge with visual mesh or texture information to solve tasks, understanding latent task requirements through emotional language interpretation, mapping spatial descriptions to target states, subtask planning for multi-step operations, logical understanding, calculations, and result derivation, among others. Figure \ref{fig:horizon_reason} presents a detailed comparison of the average episode length of overall tasks. VLABench exhibits the longest horizon among both Primitive and Composite tasks, surpassing RoboCasa Atomic and RoboCasa Composite by 27.0\% and 35.1\%, respectively. Furthermore, VLABench demonstrates significantly greater multi-step reasoning depth compared to other task sets, including sub-task numbers, inferring users’ hidden semantics, integrating visual and commonsense information, spatial reasoning, and even logical reasoning, as exemplified by solving math problems.

\noindent\textbf{One-to-Many Mapping from Task Types to Instances.}  
In VLABench, a task represents a broad category of activities designed around specific assets and the actions an agent performs. These tasks are centered on object-related themes and require diverse visual information, relevant common-sense knowledge, and rich semantic input from the user. To ensure variability, each rollout introduces different target objects and receptacles, creating unique task instances. Unlike previous benchmarks \cite{liu2024libero} where similar activities were treated as separate tasks, VLABench unifies such variations under a single task category. For example, ``placing an apple on a plate" and ``placing a pear in a box" are two tasks in most task sets, but VLABench regards them as the same task because they share the same asset theme and involve nearly identical skills. This approach focuses on the underlying theme of the asset and the agent’s actions, providing a more generalized and flexible task definition. In Figure \ref{fig:one-to-many}, we present three different rollouts from three distinct tasks. These demonstrations feature diverse language instructions, entirely different object combinations, and target objects, significantly varied visual information, and other domain randomizations, which will be elaborated further in the section \ref{sec:domain_randomization}.

\begin{table}[t]
    \centering
    \vspace{2mm}
    \renewcommand{\arraystretch}{1.27}
    \resizebox{0.5\textwidth}{!}{
    \begin{tabular}{|c|c|c||c|c|c|}
    \hline
    \textbf{Grasp Obj} & \textbf{N-Cate} & \textbf{N-Obj} & \textbf{Recep} & \textbf{N-Cate} & \textbf{N-Obj}\\ \hline
    Billiards & 2 & 24 & Billiards Table & 1 & 1 \\ \hline
    Books & 8 & 52 & Shelf & 1 & 10 \\ \hline
    Baked Goods & 5 & 60 & Microwave & 1 & 5 \\ \hline    
    Condiment & 5 & 50 & Cabinet & 1 & 5 \\ \hline
    Dessert & 4 & 58 & Tray & 1 & 15 \\ \hline
    Drink & 9 & 130 & Fridge & 1 & 5 \\ \hline
    Flower & 9 & 25 & Vase & 1 & 10 \\ \hline
    Fruit & 11 & 227 & Box Container & 2 & 10 \\ \hline
    Ingredient & 16 & 181 & Cutting Board & 1 & 15 \\ \hline
    Mahjong & 1 & 38 & Counter & 1 & 20 \\ \hline
    Number Cube & 1 & 10 & Safe & 1 & 5\\ \hline
    Painting & 1 & 286 & Stove & 1 & 5 \\ \hline
    Poker & 1 & 54 & Table & 1 & 20 \\ \hline 
    Snack & 8 & 97 & Juicer & 1 & 3\\ \hline
    Flatware & 4 & 80 & Crockery & 7 & 136\\ \hline
    Tool & 9 & 49 & Coffee Machine & 1 & 5\\ \hline
    Toy & 35 & 140 & Placemat & 1 & 10\\ \hline
    Chemistry Solution & 1 & 30 & Tube Container & 1 & 1\\ \hline
    Name Tag & 1 & 30 & Flask & 1 & 5 \\ \hline
    \end{tabular}}
    \caption{\textbf{Assets statics}. N-cate denotes the total number of object categories, while N-obj represents the total count of object instances. This table lists most of the assets.}
    \label{tab:asset_two_col}
    \vspace{-1em}
\end{table}

\subsection{Task Observation}
Each task in VLABench supports multi-view RGB-D images, semantic segmentation images, and point cloud inputs. Figure \ref{fig:task_obs} illustrates an example, showcasing the visualized point cloud data along with images from multiple viewpoints. Similar to general standard RLDS format datasets \cite{padalkar2023open}, each demonstration in VLABench not only includes the aforementioned multi-view RGB-D images and point clouds but also comprises: a list of language instructions, episode terminal, sparse reward, actions, full observations including joint positions, joint velocities, end effector position and orientation, grasping state, etc.

\noindent \subsection{Domain Randomization}
\label{sec:domain_randomization}
To ensure task diversity and broad data distribution, each task in VLABench incorporates multiple domain randomization techniques. These diversifications include:
\begin{itemize}
    \item \textbf{Mesh\&Texture Randomization.} This refers to the random variation of different instances within the same object category. For example, if a task scene requires an apple, the apple’s mesh is randomly selected from a pool of 20 distinct instances.
    \item \textbf{Position\&Orientation Randomization.} The default values for this randomized attribute are set as follows: the position offset is a random value within the range $[-0.05, 0.05]$ along the x and y directions, and the orientation is randomized with the yaw angle in the range $[-\pi/10, \pi/10]$. 
    
    In certain tasks, including \textit{SelectFruit}, grid sampling is employed for the random distribution of scene objects. Objects are constrained to be distributed within a grid space based on a maximum distance limit and are further subjected to the aforementioned basic pose offset.
    \item \textbf{Mesh Scale Randomization.} For the same mesh, VLABench scales the size of objects within a reasonable range, with the default scaling range set to $[0.95, 1.05]$.
    \item \textbf{Visual Disturbance.} VLABench employs random transformations of scenes and their relative positions, along with texture randomization of elements such as desks, floors, and walls, to achieve robust visual perturbations. In addition to the aforementioned color space transformations, the lighting intensity is randomly augmented within the range of $[0.8, 1.2]$.
    \item \textbf{Random Distractors.} VLABench requires different approaches to interpret scenes and extract key visual information accurately. To further enhance the robustness of task settings, we introduced the option to add irrelevant distractor objects to the tasks. For example, in the \textit{SelectToy} task, 1–2 fruits can be included as visual distractors.
\end{itemize}

\begin{figure}[t]
    \centering
    \includegraphics[width=8cm]{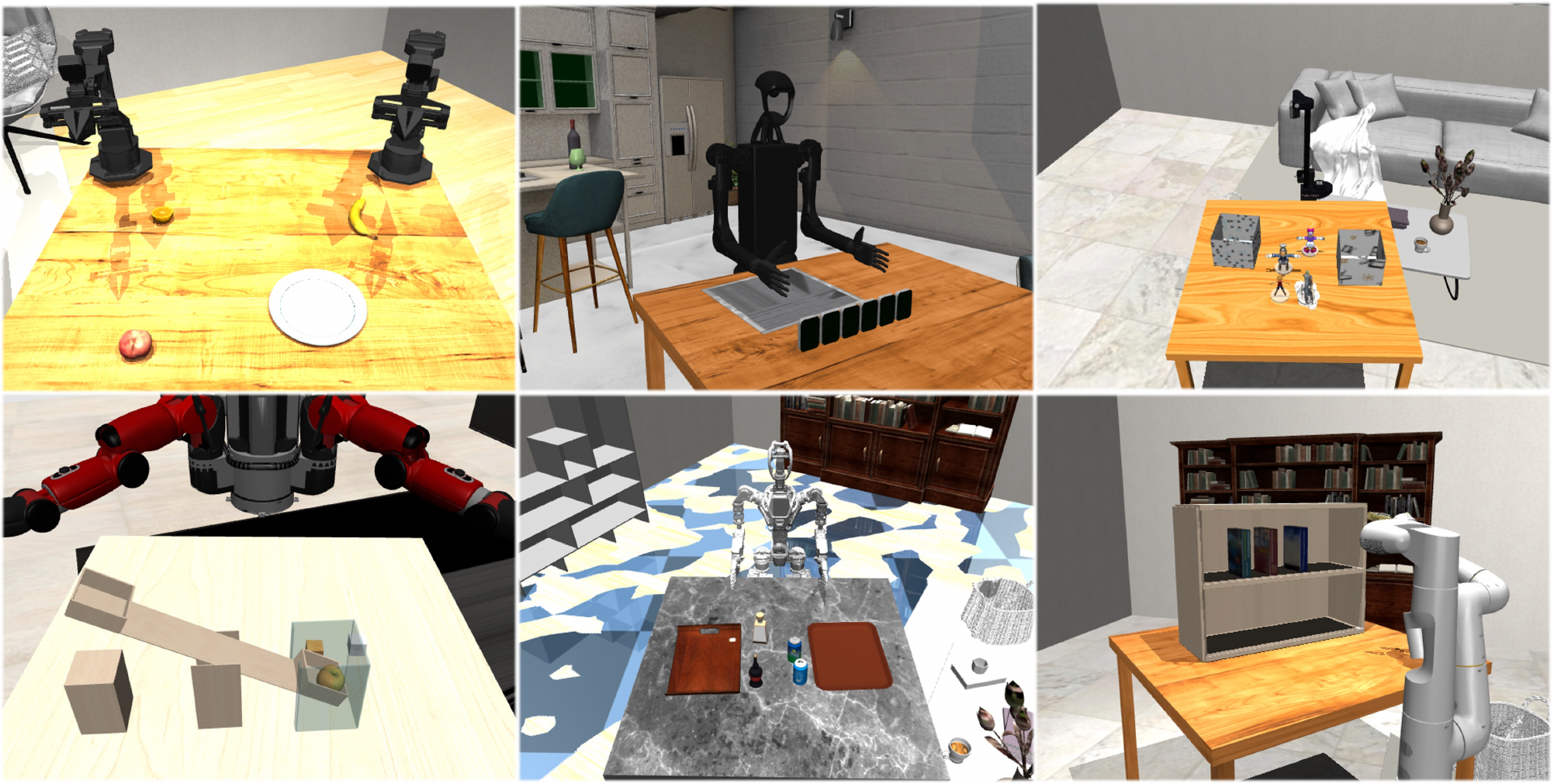}
    \caption{\textbf{Cross embodiment.} VLABench supports a wide range of different embodiments.}
    \label{fig:cross_emb}
\end{figure}

\section{Simulation and Framework}
\subsection{Scenes}
\begin{figure}[t]
    \centering
    \includegraphics[width=8cm]{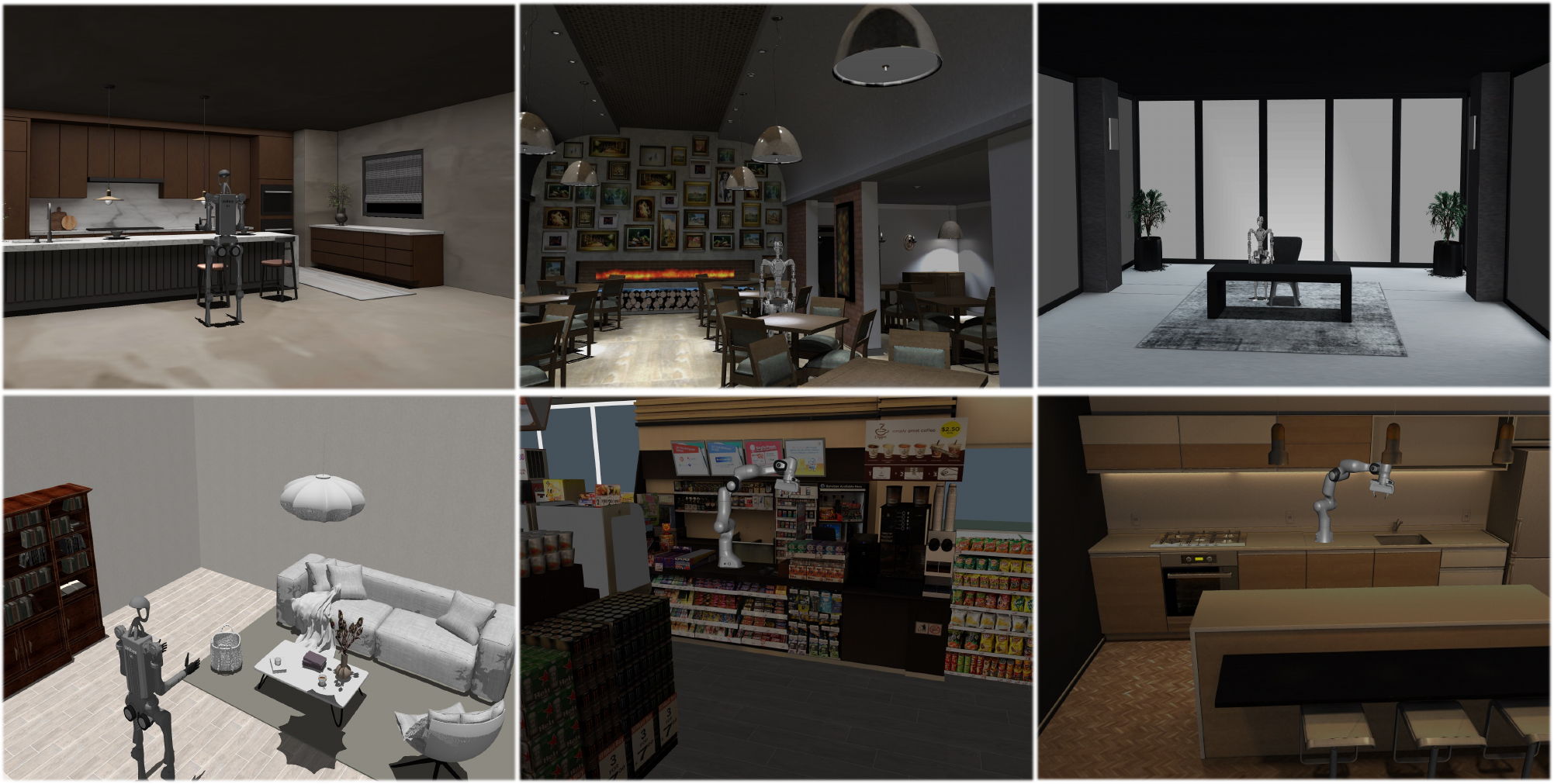}
    \caption{\textbf{Diverse scenes.} VLABench supports a wide range of different scenes.}
    \label{fig:scene}
\end{figure}

To ensure diverse task environments and rich visual inputs, we curated over 20 distinct scene types, drawing inspiration from real-life contexts and task-specific backgrounds. These scenes span everyday household settings, such as kitchens, living rooms, and dining areas, and dynamic social scenarios, including shopping malls, supermarkets, chemistry laboratories, and medical rooms. Figure \ref{fig:scene} highlights a small part of these carefully designed scenes. Beyond the variety of scene types and structures, we incorporated over 20 unique material textures for floors and walls, further enriching the visual complexity and enhancing the overall data diversity.
\subsection{Cross Embodiment}
To enable the creation of more diverse task types and datasets, VLABench supports various embodiments, including multiple models of single-arm and dual-arm robots, humanoid robots, quadrupedal robots equipped with end-effectors, and mobile robots. Figure \ref{fig:cross_emb} illustrates the performance of these different embodiments within VLABench.

\subsection{Assets}
\label{appendix:assets}
In Section \ref{sec:simulation}, we provide a brief overview of our asset library. Assets are divided into two main categories: objects-to-grasp and receptacles. For objects-to-grasp, recommended grasping points need to be annotated and are represented in the XML file using sites with \textit{class=grasppoint}. For receptacles, both bounding boxes and recommended placement points are required: the former is annotated using sites with \textit{class=keypoint}, while the latter is represented with \textit{class=placepoint}. In the coarse and large-scale pre-annotation process, we annotated grasp points on all objects-to-grasp with Graspnet \cite{fang2020graspnet, fang2023robust} and manually refined them as needed. For receptacles, we used SAM \cite{kirillov2023segment} to assist in annotating bounding boxes and assigning the placement point default above the bottom of the receptacles. Subsequent manual refinement and post-processing were applied after pre-annotation. Table \ref{tab:asset_two_col} provides an overview of the rough categories and the corresponding number of assets.

\section{Dataset Building}
\subsection{Skill Library as Domain Specific Language}
To facilitate task description and execution in robotic manipulation, we design a domain-specific language (DSL) tailored for our system. The DSL provides a structured and human-readable way to define manipulation skills, their parameters, and execution sequences. By abstracting low-level commands into high-level instructions, the DSL ensures clarity, modularity, and ease of interpretation for various tasks. The DSL consists of three primary components:
\begin{itemize}
    \item \textbf{Skills.} Atomic manipulation operations such as \textit{Pick}, \textit{Place}, \textit{Lift}, etc.
    \item \textbf{Parameters.} Arguments specify each skill's details, such as the target object, orientation, and gripper state.
    \item \textbf{Task Execute Sequence.} Sequential or hierarchical combination of skills to define a complete manipulation task.
\end{itemize}

\subsection{Data Collection Progress}
The automatic data collection process in VLABench is built upon the aforementioned DSL-encapsulated code. For each designed task, a corresponding task sequence is defined to represent the order of operations required to complete the task. For example, a case in \textit{Select Fruit} requiring the robot to pick up an apple and place it in a basket can be expressed as a DSL sequence as follows. The parameters, such as grasp pose and target position, are dynamically generated based on the simulation environment, and prior annotation information.
\begin{verbatim}
Pick("Apple", 
    {"gripper_state": "close", 
    "orientation": [np.pi, 0, 0]})
Place("Basket", 
    {"pose": [0.6, 0.4, 0.15], 
     "gripper_state": "open"})
\end{verbatim}
All tasks involve the execution of Skills using motion planning algorithms for trajectory generation. Notably, the execution of the Pick Skill requires the robotic arm to first move to a preparation position. During this process, we compute the overlap between the gripper’s point cloud and the environment’s point cloud along the trajectory from the preparation position to the grasping position. This overlap is used as a rejection sampling condition to determine an appropriate grasping direction.

\subsection{Prompt for Interactive Instruction}
We have generated a diverse set of instructions for VLABench’s dataset and evaluation tasks. These linguistically rich instructions effectively assess the ability of different models to achieve a comprehensive understanding of task scenarios. All task types, including the five categories of Primitive tasks and Composite tasks, share the following system prompt. In the system prompt, \textbf{\{object list\}} and \textbf{\{target objects\}} should be replaced with the actual objects and target objects involved in each task’s scenario. For example, in the \textit{Insert Flower} task, the object list might be [``rose'', ``tulip'', ``sunflower''], while the target objects would be [``rose'']. For each data point or evaluation task, we require the generation of ten distinct instructions, all referring to the same target object but expressed in completely different ways.

\begin{tcolorbox}[colback=green!10, colframe=green!80!black, title=System Prompt Template]
I am going to make some task instructions for a robot arm. Here are some objects:\textbf{\{object list\}}. And the target entity is \textbf{\{target objects\}}.\\ 
The target entity is the object that the robotic arm is supposed to grasp, move, or perform other operations on. Our task requirements are related to the characteristics of the target object and should also reflect everyday needs for a specific item.
\end{tcolorbox}

For tasks involving common sense and world knowledge, the prompt should additionally include the following description, emphasizing the unique characteristics of the target objects. An example for \textbf{\{Task-Specific Descriptions and Emphases\}} in \textit{Select Toy with Common Sense} task is: ``The target entity is the one that the robotic arm is supposed to grasp, move, or perform other operations on. Our task requirements are related to the characteristics of the target object. The instruction should focus on IP, rather than directly saying which toy to choose.". While the few-shot examples are:[``target\_object: Donald, instruction: 'I want a toy in the Disney series.'", ``target\_object: Goku, instruction: 'Pick a toy which belongs to the dragon ball.'"].
\begin{tcolorbox}[colback=green!10, colframe=green!80!black, title=Common Sense Template]
\textbf{\{Task-Specific Descriptions and Emphases.\}}\\
Please find the target entity's specific character which is different from other target objects and combine it into the instruction.\\
\textbf{\{Task-Specific Few-shot Examples.\}}\\
Please provide the task following the format of the above example.
Please provide ten tasks that meet the above requirements and format.
\end{tcolorbox}

For tasks requiring linguistically rich instructions, the prompt extends the system prompt by incorporating the following semantic prompt. The few-shot examples in \textit{Select Toy Semantic} may be like: [``target\_object: batman, instruction: `I'm a big fan of DC series, please help me choose a suitable toy.'", ``target\_object: Luffy, instruction: `Today is my friend’s birthday, and I want to buy a Luffy figure for him. Could you help me wrap it? Thank you!'"].

\begin{tcolorbox}[colback=green!10, colframe=green!80!black, title=Semantic Template]
Please find the target entity's specific character which is different from other target objects and combine it into the instruction.
Do not directly mention the target entity by name and avoid explicitly stating the need for the object. Instead, create tasks that reflect real-life scenarios where the need for the object is implied through casual, everyday observations. The task should suggest a need without saying it directly, focusing on natural, implied requests.  
\textbf{\{Task-Specific Few-shot Examples.\}}\\
Please provide the task following the format of the above example.\\
The target\_entity must be the target entity. Please provide ten tasks that meet the above requirements and format.
\end{tcolorbox}

Composite tasks integrate the abilities and skills involved in primitive tasks, with each composite task featuring its unique scenario and context. In this setup, while the system prompt remains shared, each task is accompanied by a specific prompt. Here, we present the specific prompt for the \textit{Cluster Book} task.
\begin{tcolorbox}[colback=green!10, colframe=green!80!black, title=Composite Task Example: Cluster Book]
The task now is to classify the books. Please design real-life scenarios where there is a need to categorize books and generate instruction based on the classification requirement. \\
You cannot specify the exact classification method; just create a realistic scenario that requires classification and instruct it to categorize the books in front of it.\\
Please make the generated instructions more diverse in terms of conversational language, tone, and scenarios. Avoid sticking to a single-sentence structure.\\
\textbf{\{Task-Specific Few-shot Examples.\}}\\
Please provide the task following the format of the above example.\\
Please provide ten tasks that meet the above requirements and format.
\end{tcolorbox}

\section{Experiment Implementation}
\subsection{VLA Setting}
To assess the generalization ability of various VLAs, we primarily fine-tune OpenVLA, Octo, and RDT-1B using our dataset. We utilize the original open-source code and adhere to the default hyperparameters set by the authors. To ensure comparability across datasets of varying sizes, we fix the number of training epochs instead of the maximum training steps. Given that OpenVLA has 7B parameters, we apply the recommended LoRA strategy in all experiments, rather than performing full parameter fine-tuning. In contrast, the other two models undergo full parameter fine-tuning. We train all models until convergence is achieved. Notably, Octo exhibits a certain reluctance to converge, which might be attributed to its relatively low level of generality. Consequently, we conduct training for over 5 epochs to obtain the optimal fit. Note that we adhere to the default configurations of these models: OpenVLA and Octo process a single-view image as input, whereas RDT-1B utilizes three different views. All experiments are conducted on 4 NVIDIA A800 with 80GB of memory.

\subsection{Evaluation of Worksflows}
In evaluating the foundation model-based workflow algorithms, we adopt the same evaluation process and metrics used for assessing the VLAs. The procedure for evaluating each task individually is outlined as follows:

\begin{enumerate}
\item \textbf{Run the Base Model Workflow.} Execute the base workflow in the specified environment and record the corresponding outputs, with particular emphasis on data related to the model's target entity detection information.

\item \textbf{Task Evaluation.} Once the relevant information has been collected, the success of the task and the accuracy of target identification are assessed. Specifically, correct identification of a target contributes 20\% of the total score, while task success will award full points.

\item \textbf{Final Score Calculation.} After evaluating individual tasks several times, the scores for each time task are aggregated to yield the final score for the model under each configuration.
\end{enumerate}

By applying this evaluation framework, we ensure a consistent and comprehensive assessment of the model's performance across different tasks and settings.



\subsection{Evaluation of VLMs}
\label{appendix:eval vlm}
As discussed in Section \ref{sec:eval vlm}, the entire evaluation process of VLMs can be simplified to DSL generation and the score can be computed through direct graph matching. The assessment of the skill sequences output by the VLM is based on the following four metrics.


\noindent\textbf{Skill Recall Rate (SR).} 
    We use SR as the coarsest-grained metric to evaluate the model's capability to identify and invoke the correct skills.
    \begin{equation}
        SR = \frac{|\textit{SL}_{\text{gt}} \cap \textit{SL}_{\text{pred}}|}{|\textit{SL}_{\text{gt}}|}
    \end{equation}
     where $\textit{SL}_{\text{gt}}$ represents the list of skills manually labeled for completing tasks, and $\textit{SL}_{\text{pred}}$ refers to the list of skills predicted by the model. The denominator corresponds to the total number of relevant skills in the dataset, while the numerator counts the intersection of the relevant skills and those correctly identified by the model.

\noindent\textbf{Parameter Recall Rate (PR).}
     The PR quantifies the model's ability to correctly identify the parameters associated with each skill. In many cases, each skill is contingent upon specific parameters, which are often represented by the labels of relevant objects within an image. The PR thus measures the model's accuracy in recognizing and interpreting these parameters, a crucial aspect for ensuring the correct execution of the task. Accurate parameter identification is fundamental not only for skill invocation but also for the model's overall performance in real-world applications. A higher PR indicates a higher accuracy of the parameters predicted by the model, thus ensuring that the model correctly identifies the entities that need to be valued in the figure.
    \begin{equation}
        PR = \frac{|\textit{Param}_{\text{gt}} \cap \textit{Param}_{\text{pred}}|}{|\textit{Param}_{\text{gt}}|}
    \end{equation}
    where $\textit{Param}_{\text{gt}}$ refers to the list of parameters manually labeled for each skill, and $\textit{Param}_{\text{pred}}$  denotes the list of parameters predicted by the model. 

\noindent\textbf{Skill\&Parameter Recall Rate (SPR).} 
  Unlike the individual metrics SR and PR, SPR requires the model to identify both the correct skills and the exact parameters associated with each skill. It provides a more comprehensive and strict evaluation of the model's ability of scene understanding and task planning in a real-world context. This metric is particularly useful in evaluating scenarios where both skills and their contextual parameters are critical for task execution, such as in visual recognition tasks where precise associations between actions and objects are necessary.
    \begin{equation}
        SPR = \frac{|\textit{SP-Pair}_{\text{gt}} \cap \textit{SP-Pair}_{\text{pred}}|}{|\textit{SP-Pair}_{\text{gt}}|}
    \end{equation}
    where $\textit{SP-Pair}_{\text{gt}}$ represents the set of all manually labeled skill-parameter combinations, and $\textit{SP-Pair}_{\text{pred}}$ refers to the corresponding combinations predicted by the model. 
    
    \paragraph{Precise Matching Rate (PM).} 
    In addition to evaluating the correctness of skill-parameter matching, PM places greater emphasis on assessing the logical dependencies of the skill sequence, particularly for tasks with strict temporal requirements. Instead of totally strict sequential order, this metric focuses on ensuring that the necessary dependencies are satisfied for successful task execution. For example, in \textit{Make Juice} task, the model must ensure that the juicer is opened before adding fruit, but the order of adding apples versus oranges is irrelevant.
    
    We begin by aggregating the skill sequence according to predefined operational patterns and constructing a directed acyclic graph (DAG) with a designated source node to represent the logical dependencies among operations. A match is defined as a node in the model-generated graph that shares the same skill name and parameters as a corresponding node in the ground-truth graph while also satisfying the logical dependency relationships, e.g. incoming and outgoing edges.The formula for this metric is as follows:
    \begin{equation}
        PM = \frac{|\textit{Node}_{\text{matched}}|}{|\textit{Node}_{\text{total}}|}
    \end{equation}
    where the numerator $\textit{Node}_{\text{matched}}$ represents the number of nodes in the model-generated DAG that match the corresponding nodes in the ground-truth DAG.   $\textit{Node}_{\text{total}}$ represents the total number of nodes in the ground-truth DAG.

Finally, these four scores are combined using predetermined weights to compute a total score for each model. The formula for this metric is as follows:
\begin{equation}
Score = w_1 \cdot SR + w_2 \cdot PR + w_3 \cdot SPR + w_4 \cdot PM
\label{eq:overall_score}
\end{equation}
where $w_1, w_2, w_3, w_4$ are the weights of different metrics, with the constraint $w_1 + w_2 + w_3 + w_4 = 1$.

\section{Detailed Analysis and Case Study}
\begin{figure}[t]
    \centering
    \includegraphics[width=\linewidth]{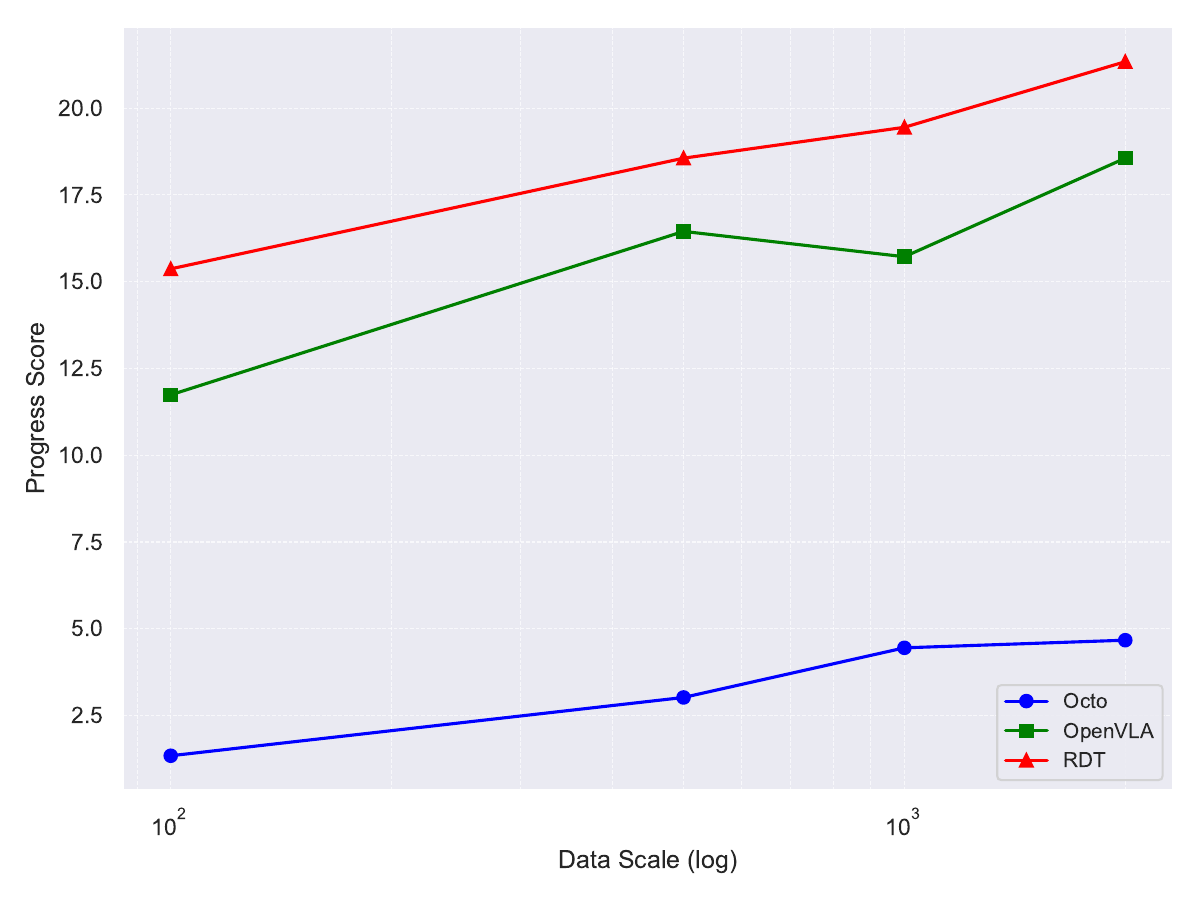}
    \caption{\textbf{Scaling trend.} This result is evaluated on \textit{Put Box on Paintig Task} with data scales of 100, 500, 1000, and 2000.}
    \vspace{-1em}
    \label{fig:scaling}
\end{figure}
\begin{table}[t]
    \centering
    \vspace{2mm}
    \begin{minipage}[t]{0.48\textwidth}
        \centering
        \renewcommand{\arraystretch}{1.25}
        \begin{tabular}{p{2cm}cc}
            \hline
            \textbf{Model} & \textbf{From Scratch} & \textbf{From Pretrained} \\ \hline
            Octo & 1.02 & 1.34  \\
            OpenVLA & 3.02 & 11.74 \\
            RDT-1B & 6.26 & 15.37 \\
            \hline 
        \end{tabular}
        \caption{Ablation of fine-tuning from scratch and pretrain. Evaluated on primitive tasks with seen objects.}
        \label{subtable:pretrain_vs_scratch}
    \end{minipage}
    \hfill
    \begin{minipage}[t]{0.48\textwidth}
        \centering
        \vspace{5.5mm}
        \renewcommand{\arraystretch}{1.25}
        \begin{tabular}{p{4cm}c}
            \hline
            \textbf{Model} & \textbf{Avg PS} \\ \hline
            RDT-1B\_open\_step\_64 & 15.37 \\
            RDT-1B\_open\_step\_32 & 15.52 \\
            RDT-1B\_cls & 17.68 \\ \hline
        \end{tabular}
        \caption{Comparison of open-loop and closed-loop control for RDT-1B. Evaluated on primitive tasks with seen objects.}
        \label{tab:open_vs_close}
    \end{minipage}
    \vspace{-1em}
\end{table}

\subsection{Ablations and Analysis for VLAs}
\label{appendix:ablation vla}
Experimental results show that the current open-source VLAs perform poorly on our tasks. On one hand, this can be attributed to the high difficulty of VLABench tasks, which impose stringent requirements on the generalization capabilities of the models. More importantly, the limitations and deficiencies in both the architecture and pretraining process of current VLAs make it challenging for them to adapt effectively to downstream tasks after large-scale pretraining, especially under fine-tuning scenarios with diverse data distributions. This stands in stark contrast to LLMs, which excel in adapting to downstream tasks with minimal fine-tuning on small datasets.
\begin{figure*}[h!]
    \centering
    \begin{minipage}[t]{0.48\textwidth}
        \vspace{0pt}

        \centering
        \begin{subfigure}[b]{0.48\textwidth}
            \centering
            \includegraphics[width=\textwidth]{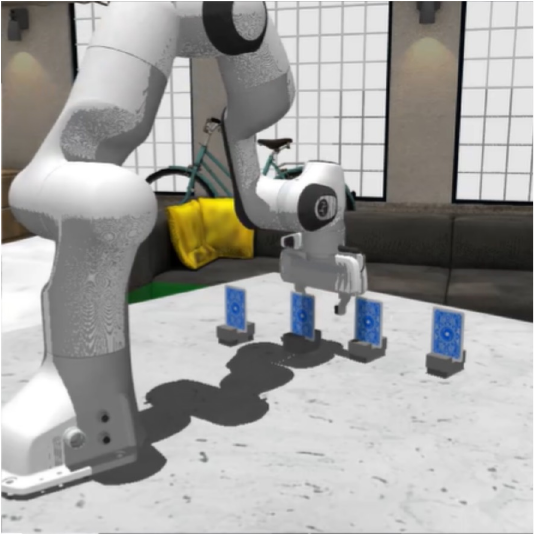}
            \caption{Select Poker}
            \label{fig:subfigA}
        \end{subfigure}
        \hfill
        \begin{subfigure}[b]{0.48\textwidth}
            \centering
            \includegraphics[width=\textwidth]{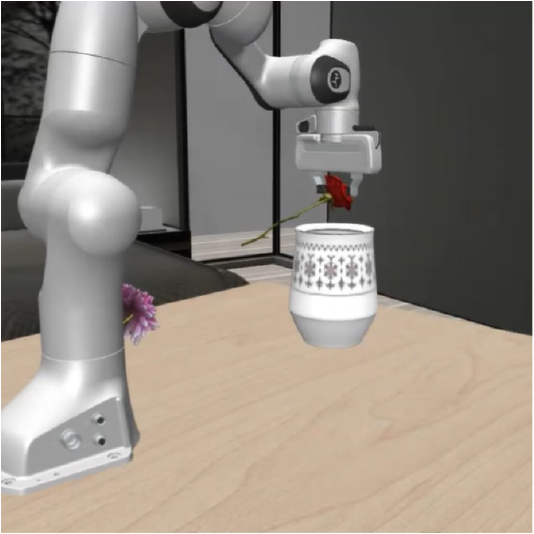}
            \caption{Insert Flower}
            \label{fig:subfigB}
        \end{subfigure}
        
        \vspace{5pt}
        \begin{subfigure}[b]{0.48\textwidth}
            \centering
            \includegraphics[width=\textwidth]{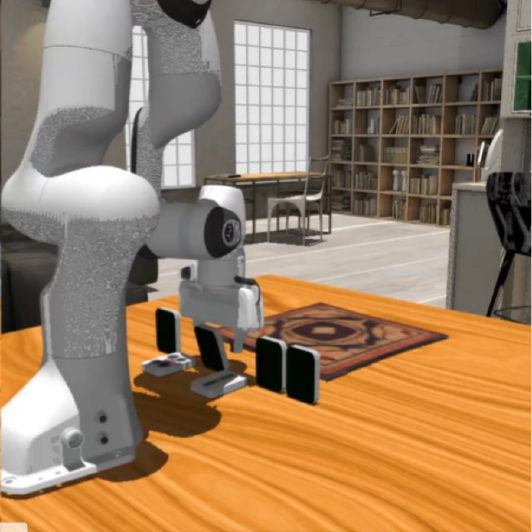}
            \caption{Select Mahjong}
            \label{fig:subfigC}
        \end{subfigure}
        \hfill
        \begin{subfigure}[b]{0.48\textwidth}
            \centering
            \includegraphics[width=\textwidth]{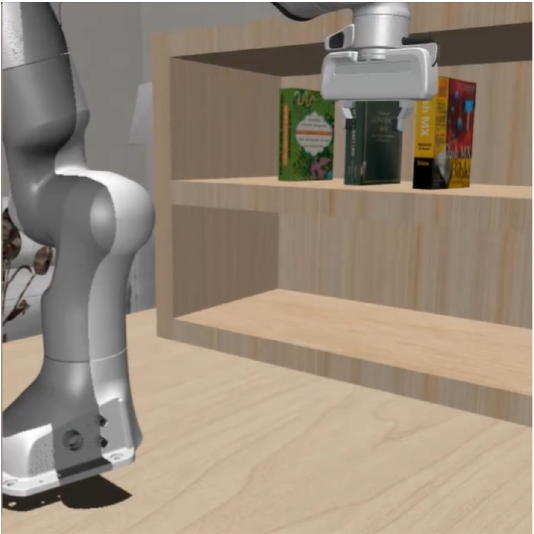}
            \caption{Select Book}
            \label{fig:subfigD}
        \end{subfigure}
        \caption{Failure case: failures caused by the inability of the algorithm to percept task scene and plan on rotation.}
        \label{fig:rotation_failure}
    \end{minipage}
    \hfill
    \begin{minipage}[t]{0.48\textwidth}
    \vspace{0pt}

        \centering
        \begin{subfigure}[t]{0.48\textwidth}
            \centering
            \includegraphics[width=\textwidth]{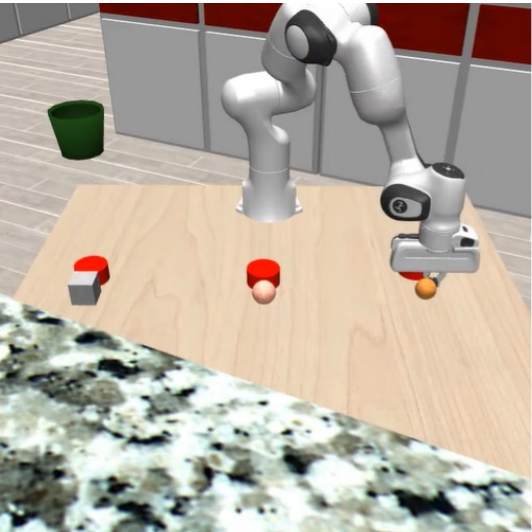}
            \caption{Physics QA}
            \label{fig:subfigA_grasp}
        \end{subfigure}
        \hfill
        \begin{subfigure}[t]{0.48\textwidth}
            \centering
            \includegraphics[width=\textwidth]{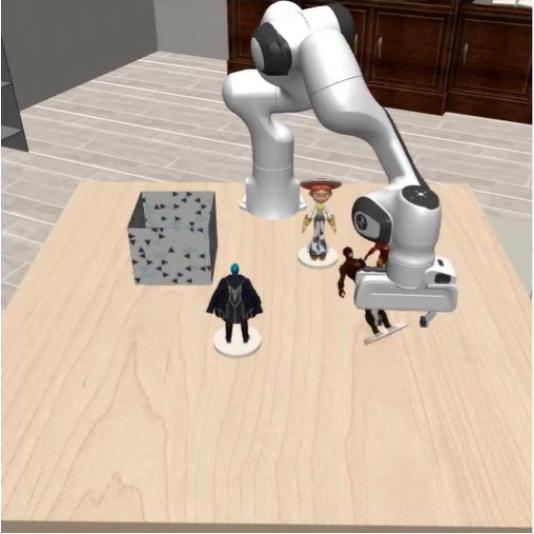}
            \caption{Select Toy}
            \label{fig:subfigB_grasp}
        \end{subfigure}
        \vspace{-9pt}
        \caption{Failure case: fails to perceive object interactions.}
        \label{fig:grasp_failure}
        \vfill
        \begin{subfigure}[b]{0.48\textwidth}
            \centering
            \includegraphics[width=\textwidth]{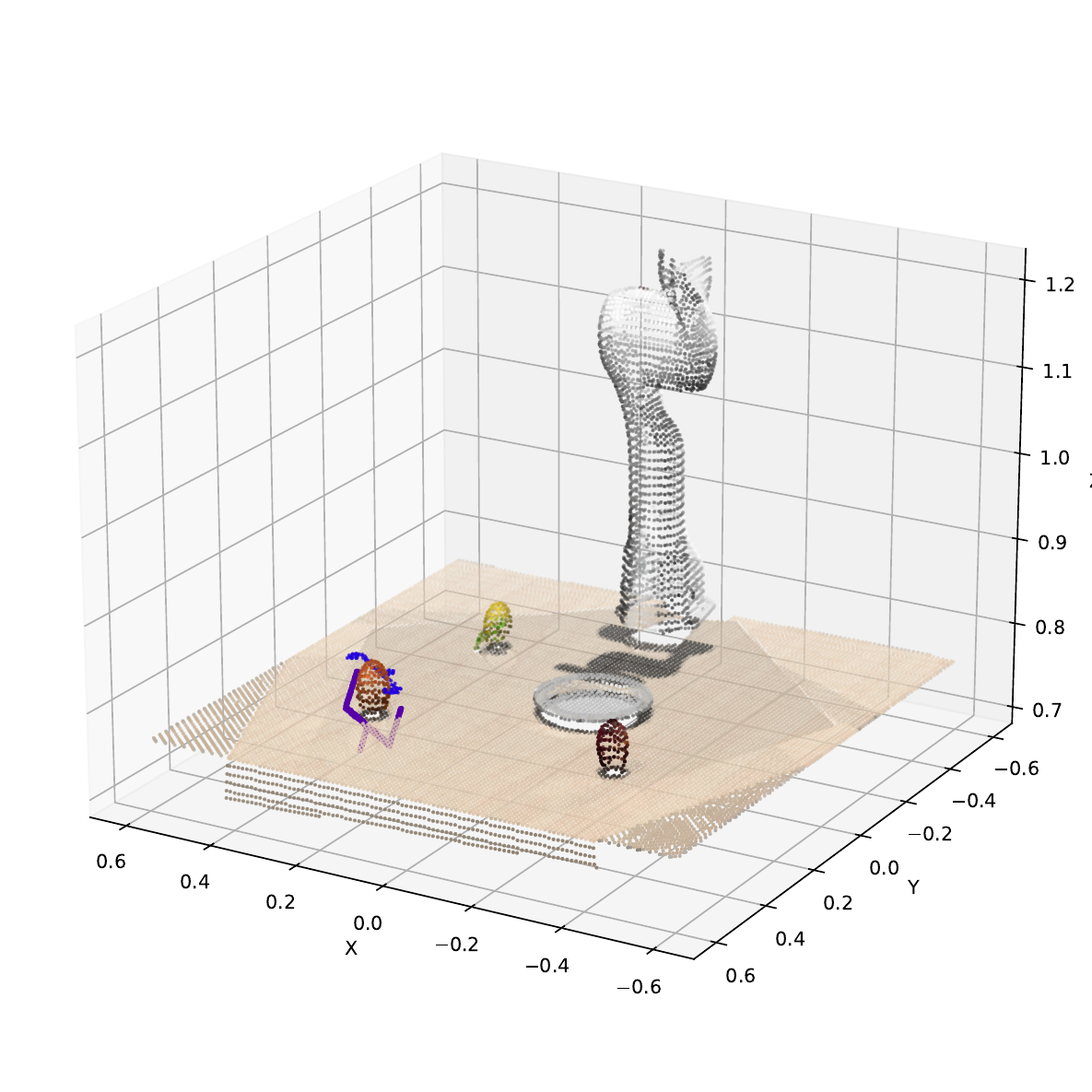}
            \caption{Graspnet successes.}
            \label{fig:subfigA_grasp}
        \end{subfigure}
        \hfill
        \begin{subfigure}[b]{0.45\textwidth}
            \centering
            \includegraphics[width=\textwidth]{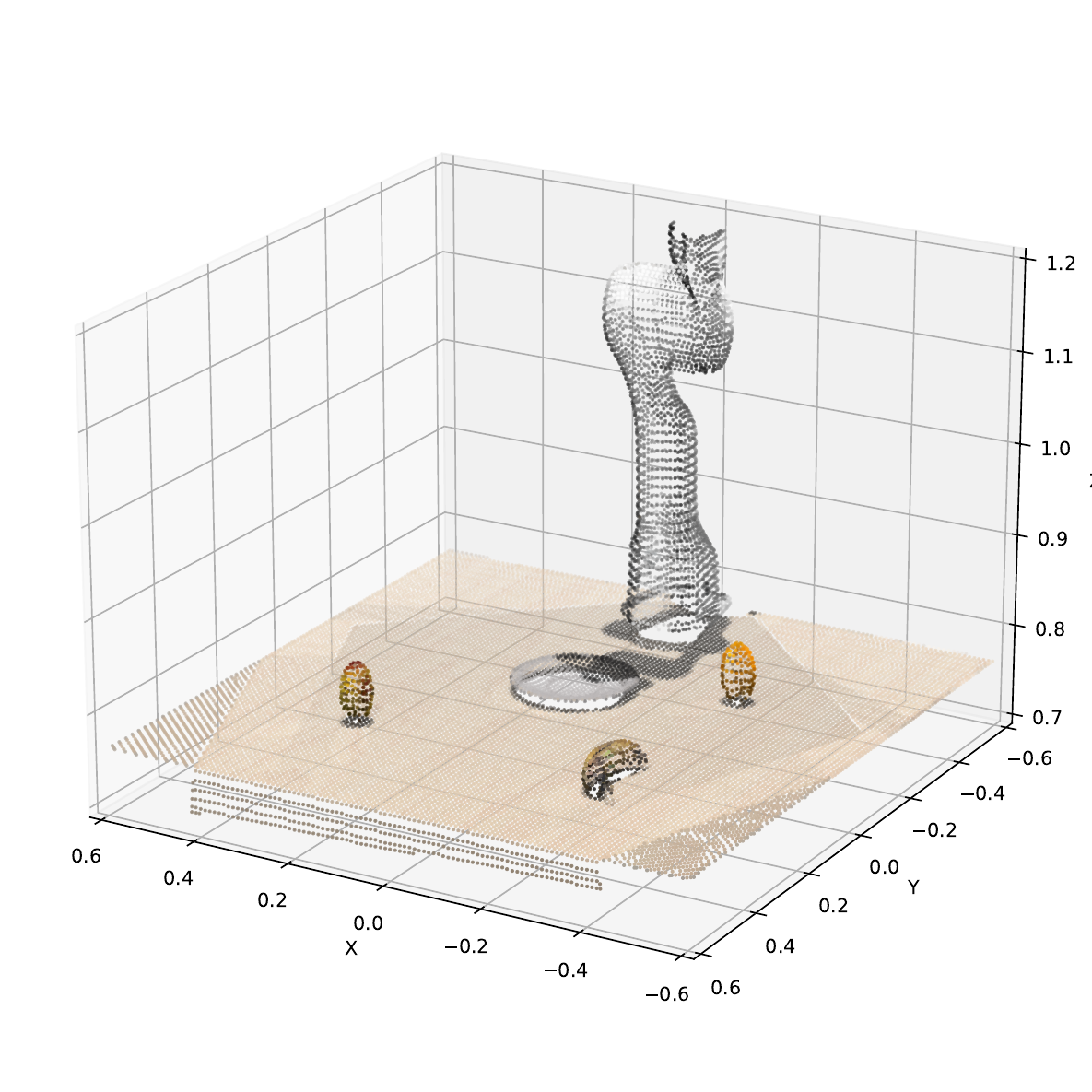}
            \caption{Graspnet fails.}
            \label{fig:subfigB_grasp}
        \end{subfigure}
        \vspace{-8pt}
        \caption{Failure case: Graspnet fails in generating valid grasp points, resulting in task failing.}
        \label{fig:find_grasp_failure}
        \par\vspace{0pt}
        \vfill
       
    \end{minipage}
\end{figure*}
To further illustrate the aforementioned issues, we conducted several ablation experiments:
\begin{itemize}
    \item \textbf{Data Scaling.} More data implies a greater number of visual-language to trajectory mappings. For specific primitive tasks, we expanded the dataset to 2,000 samples and conducted separate evaluations on three models using datasets of varying scales: 100, 500, 1,000, and 2,000 samples. However, the experimental results show that under diverse data distributions, the task success rates of all three models remain consistently low across the four scales. This issue is primarily reflected in their operational accuracy. As illustrated in Figure \ref{fig:scaling}, while the trajectory generation becomes increasingly smooth and the PS shows a slight improvement with larger datasets, the overall task success rate remains notably low.
    \item \textbf{Pretrained Effect.} We also conducted an evaluation of models trained with the fine-tuning dataset from scratch, with the results summarized in Table \ref{subtable:pretrain_vs_scratch}. The findings indicate that models of this scale struggle to quickly adapt to downstream tasks with limited data. It is reasonable to infer that pretraining on large-scale, domain-relevant data can significantly facilitate the faster transfer of VLAs to downstream tasks.
    \item \textbf{Closed-loop for RDT.} Following the open-source RDT framework, the primary experiment employs a single-trajectory inference scheme with 64 trajectory points, implemented using open-loop control. However, open-loop control is prone to error accumulation. To address this, we conducted additional evaluations of RDT using closed-loop control. The results in Table \ref{tab:open_vs_close} show that closed-loop control achieves slightly better performance compared to open-loop control. This suggests that the low success rate in task execution is primarily due to the inherent limitations of the model itself.
\end{itemize}

To analyze why these models perform poorly, we base our discussion on experimental results and observations from two key perspectives.

\noindent\textbf{Limitation of Model Architecture.} 

\begin{itemize}
    \item \textbf{Incomplete Information Intake}. Some shortcomings in the model architecture result in this issue. For instance, OpenVLA and Octo only process single images with a resolution of 224×224, which inherently puts them at a disadvantage when the input images contain occlusions or require finer texture details. Similarly, due to issues with perspective and low resolution, directly mapping visual information to precise spatial coordinate points becomes challenging.
    \item \textbf{Lack of Memory}. Current models only accept inputs representing the current state, lacking position embeddings to capture temporal sequences or tokens to represent historical actions. This limitation can cause the model’s behavior to become stuck in certain states. This issue is particularly pronounced in long-horizon tasks, where the model may ``forget" previous actions and repeatedly perform the same behavior.
    \item \textbf{Inherent Flaws of Different Architecture.} The VLAs we used primarily include two forms: transformer-based next-token prediction architectures and diffusion model-based architectures. The former, leveraging VLMs, benefits from pretraining on world knowledge but inherently suffers from precision loss due to the discretization required by action tokenization. On the other hand, diffusion policies are better suited for continuous spatial distributions, yet they lack visual and language pretraining. Additionally, diffusion models rely on multiple large encoders, such as T5, making it challenging to jointly fine-tune parameters during unified training. This limitation contributes to the poor performance of diffusion policies in VLABench tasks requiring common sense.
\end{itemize}

\noindent\textbf{Shortcomings of Pretrain.}
VLA pretraining has been proven effective for efficient transfer to downstream tasks. However, the current pretraining approaches may have certain issues. For example, RT-2 \cite{rt22023arxiv} highlights a pretraining strategy that jointly trains on multiple text tasks and text-visual tasks to preserve the model’s inherent language and reasoning capabilities, resulting in impressive generalization behaviors. In contrast, OpenVLA, which is also based on VLM, is pretrained solely on trajectory datasets. This likely leads to the degradation of VLM’s original capabilities, such as commonsense knowledge and reasoning skills.

Additionally, constrained by the availability of datasets, the scale of current VLA pretraining data is far smaller than that of language models. Drawing inspiration from the scaling laws and emergent behaviors observed in language models, there is likely a critical point and correlation between model parameter size and data volume. The scaling curve for VLA pretraining, however, remains an open topic for future research.

\subsection{Further Analysis for Workflows}
\label{appendix:ablation agent}
From the experimental results, we observe that while the framework algorithm based on the foundation model demonstrates some degree of robustness in handling complex semantic settings, the overall success rate and PS score remain relatively low. A comprehensive analysis of the failure cases reveals that the underlying issues can be broadly categorized into the following groups.

\noindent\textbf{Perception.}
One of the primary challenges lies in the model's image and spatial perception capabilities. As Voxposer is implemented as a purely text-based framework, its perception module relies directly on the ground-truth labels of all items, which are provided as input for selection. While this leverages the comprehension and generalization capabilities of large language models to understand tasks, it exposes significant limitations in scenarios that require spatial perception and image-based reasoning. Specifically, Voxposer demonstrates clear incompetence in handling tasks involving spatial awareness or detailed image descriptions.

To address this, we augmented our experimental setup by incorporating an image perception module into Voxposer. Although this adjustment improved success rates on spatial perception tasks, the overall performance deteriorated due to errors introduced by the visual perception module. A similar issue was observed in CoPA, where the SoM family of models exhibited high sensitivity to segmentation parameters, requiring extensive tuning to achieve accurate entity recognition. Even with optimization, a substantial number of incorrect object recognition cases persisted, highlighting fundamental challenges in the perception component.

\noindent\textbf{Planning.}
Another significant limitation emerges in the model's planning capabilities. After selecting the target object, the model's lack of spatial perception often prevents it from recognizing the need to adjust its pose, such as rotating the robotic arm when grasping certain objects. This deficiency leads to frequent task failures, particularly in scenarios involving objects like cardboard sheets or books, as illustrated in Figure \ref{fig:rotation_failure}. Additionally, the simple point-cloud-based center-of-mass grasping strategy employed by the model exhibits a high probability of failure when interacting with objects of complex shapes, such as toys, as shown in Figure \ref{fig:grasp_failure}.

For CoPA, similar challenges were encountered in the graspnet module, where planning grasping actions was hindered by its instability. In many instances, the module failed to identify a valid grasping point, resulting in task failures, as depicted in Figure \ref{fig:find_grasp_failure}. These issues underscore the model's inability to effectively plan and execute tasks involving diverse and irregularly shaped objects.

\begin{figure*}[t]
    \centering
    \begin{minipage}[T]{0.48\textwidth}
        \centering
        \begin{subfigure}[b]{0.48\textwidth}
            \centering
            \includegraphics[width=\textwidth]{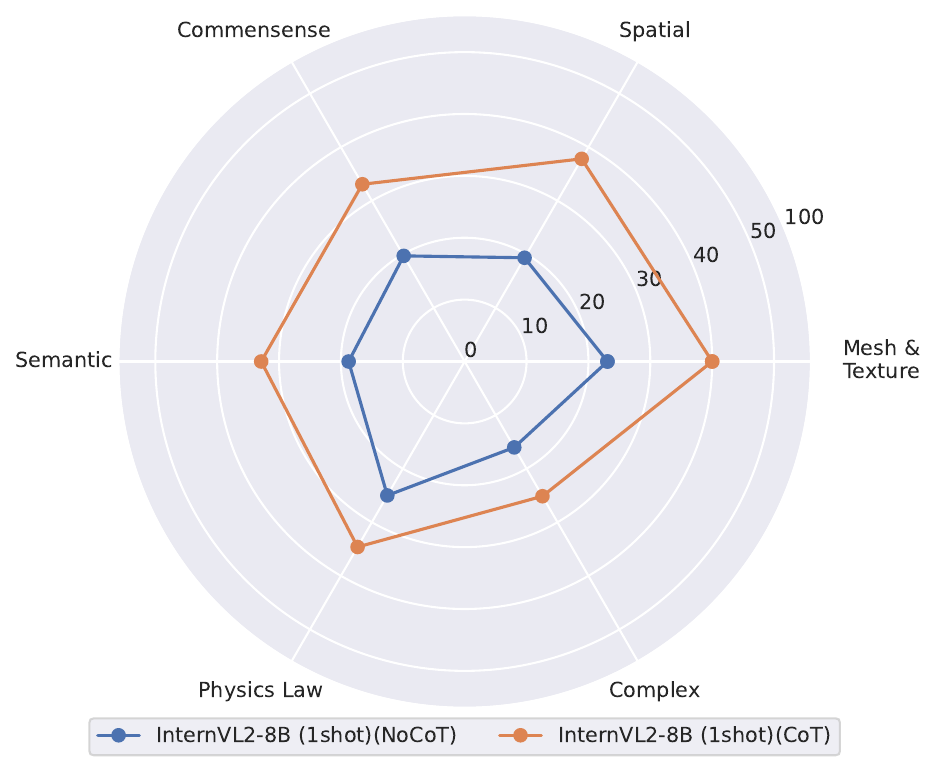}
            \caption{Result of InternVL2-8B.}
            \label{fig:subfigA}
        \end{subfigure}
        \hfill
        \begin{subfigure}[b]{0.48\textwidth}
            \centering
            \includegraphics[width=\textwidth]{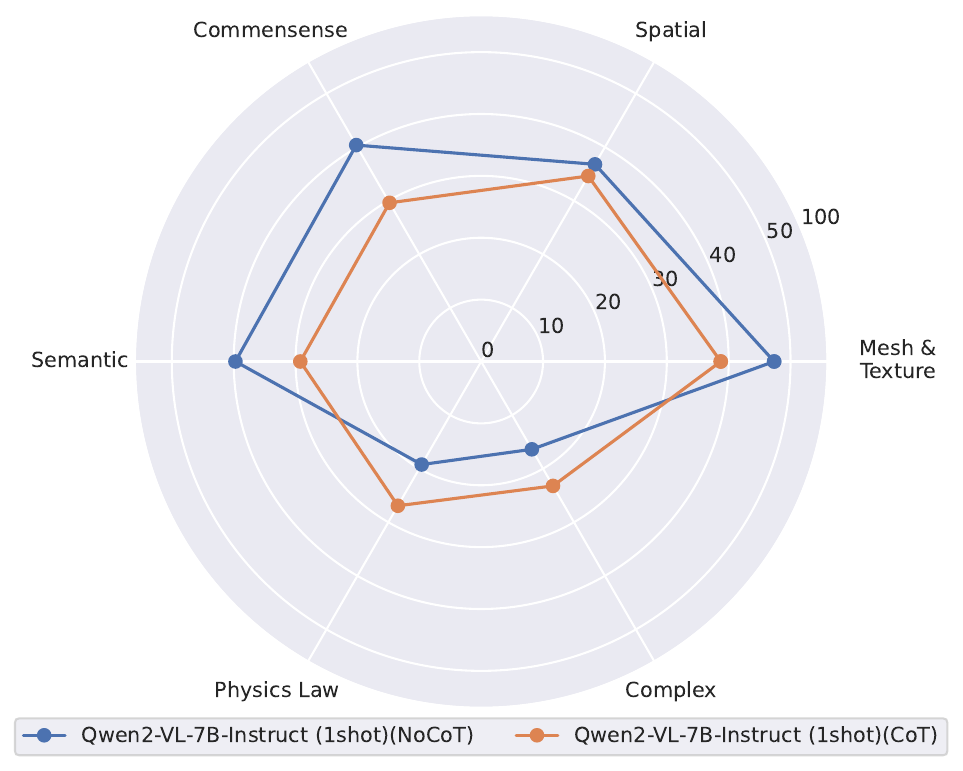}
            \caption{Result of Qwen2-VL-7B.}
            \label{fig:subfigB}
        \end{subfigure}

        \vspace{0.5cm}

        \begin{subfigure}[b]{0.48\textwidth}
            \centering
            \includegraphics[width=\textwidth]{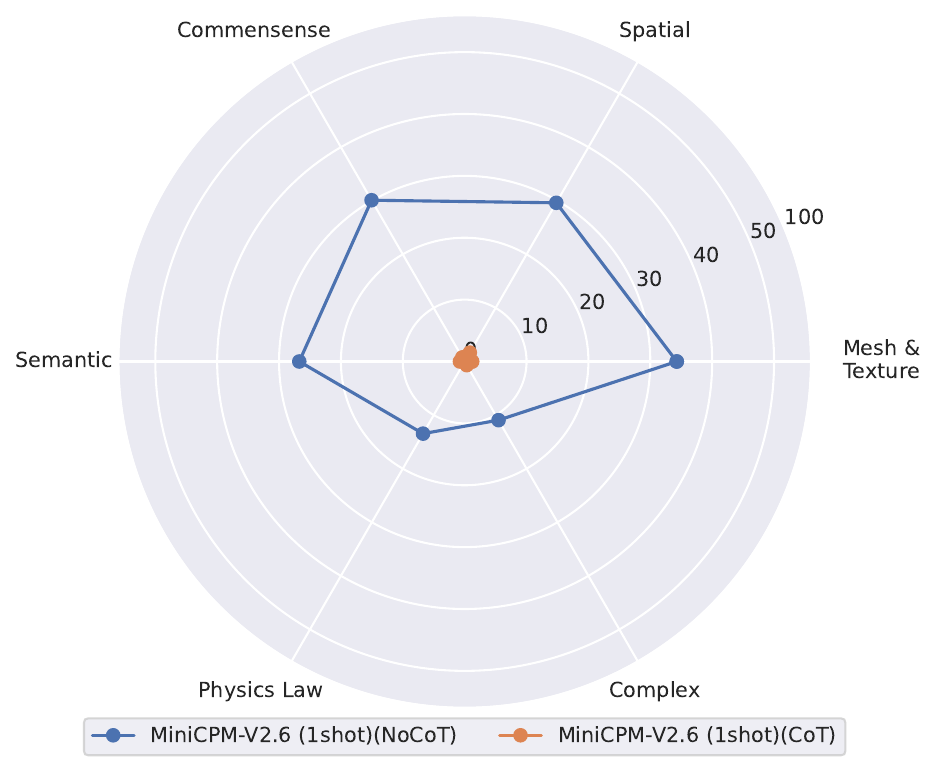}
            \caption{Result of MiniCPM-V2.6.}
            \label{fig:subfigC}
        \end{subfigure}
        \hfill
        \begin{subfigure}[b]{0.48\textwidth}
            \centering
            \includegraphics[width=\textwidth]{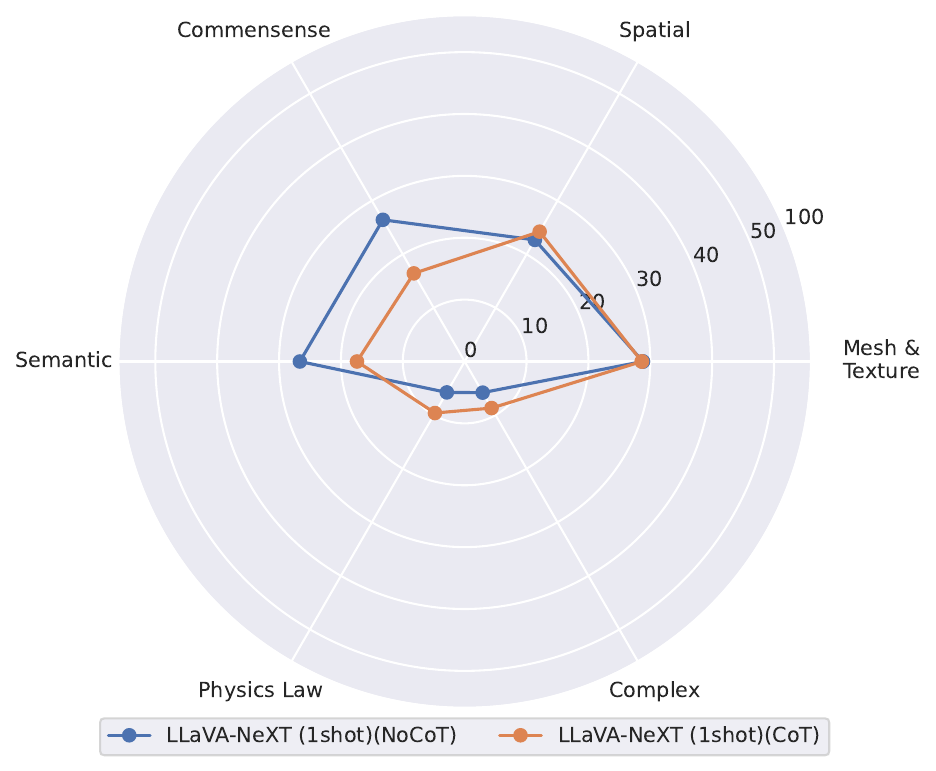}
            \caption{Result of LLaVA-NeXT.}
            \label{fig:subfigD}
        \end{subfigure}

        \caption{Variation of the six-dimensional scores of the different models in the CoT case, where the \textcolor{orange}{orange} line represents the case with CoT, and the \textcolor{blue}{blue} line represents the case without CoT.}
        \label{fig:CoT}
    \end{minipage}
    \hfill
     \begin{minipage}[T]{0.48\textwidth}
        \setlength{\abovecaptionskip}{0.cm}
        \centering
        \includegraphics[width=\linewidth]{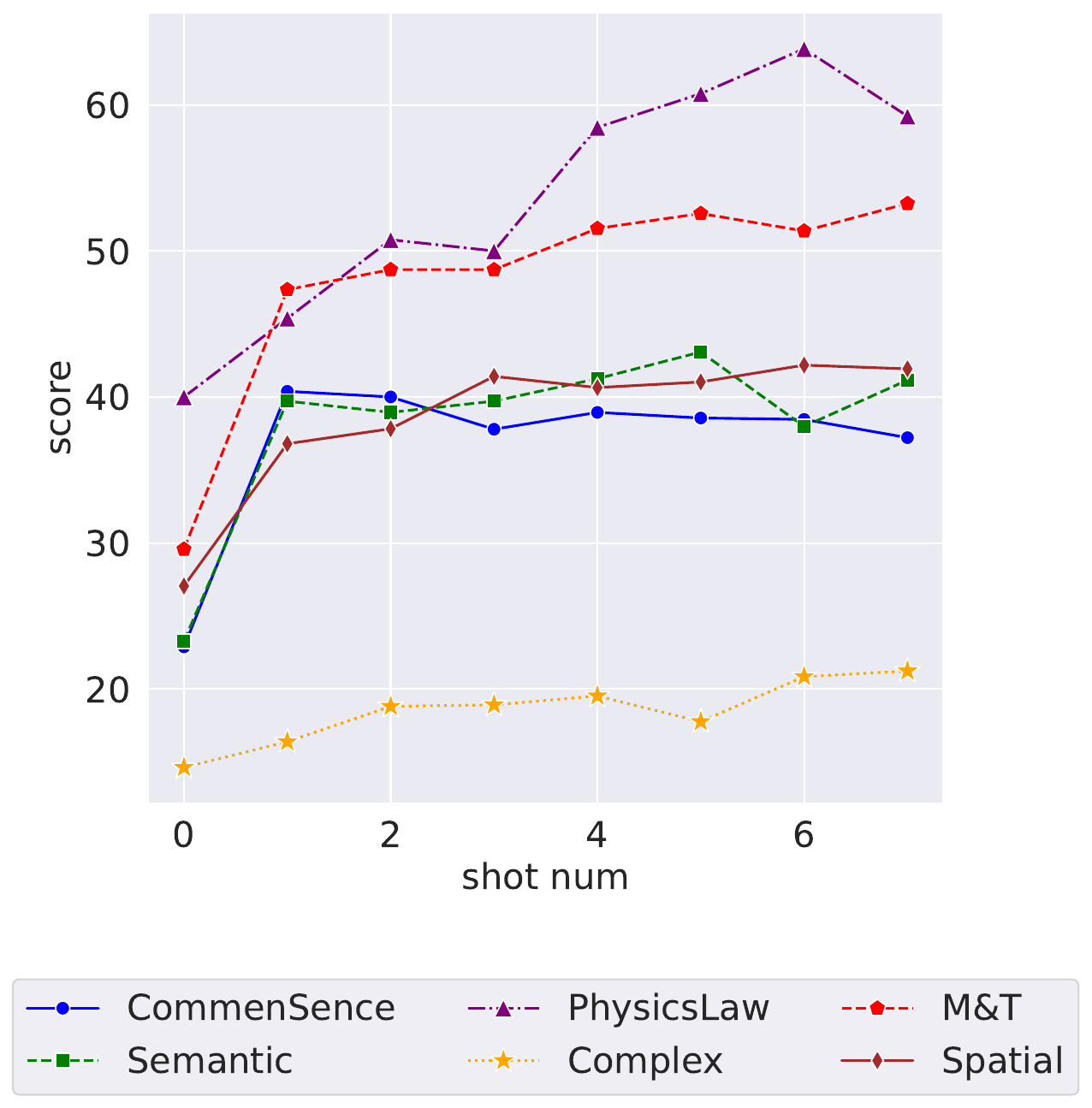}
        \caption{The impact of different few-shot settings on the performance of VLMs. As the number of few-shot examples increases, the generation quality of the model improves progressively.}
        \label{fig:few_shots}
        \par\vspace{0pt}
        \vfill
    \end{minipage}
\end{figure*}

\noindent\textbf{Module Connections.}
As hierarchical systems, such algorithms rely on the integration of multiple independent modules, which inevitably introduces errors at the interfaces between components. For example, the large language model may generate incorrect outputs, such as failing to locate the corresponding object or the constraints generated by the system may not be successfully converted into waypoints by the solver. These errors significantly reduce the system's robustness when handling diverse task conditions. The inability to reliably bridge constraints and waypoints highlights a critical limitation in the framework's modular connectivity, further undermining its ability to adapt to varying operational scenarios.

\subsection{Ablations and Analysis for VLMs}
\label{appendix:ablation vlm}
In our evaluation of VLMs, we conducted two key experiments to explore the impact of Chain-of-Thought (CoT) prompting and few-shot learning on model performance.

\noindent\textbf{Effect of CoT Prompting.}
Our investigation into the use of CoT prompting revealed a notable improvement in overall performance for the InternVL2 model, as shown in Figure \ref{fig:CoT}. Similarly, LLaVA-NeXT and Qwen2-VL demonstrated enhanced performance in challenging tasks, particularly those requiring reasoning about complex scenarios and physics laws. However, their performance on semantically common-sense tasks remained stagnant or experienced minor degradation. In contrast, the MiniCPM model exhibited significant limitations: it failed to output answers at the conclusion of the reasoning process when CoT was applied, resulting in all scores dropping to 0.0.

\noindent\textbf{Effect of Few-Shot Learning.}
As shown in Figure \ref{fig:few_shots} our exploration of few-shot learning with the Qwen2-VL model indicated that increasing the number of few-shot examples (0 to 7) enhances the model’s multimodal reasoning capabilities, particularly under CoT prompting. This enhancement was observed across both basic and complex scenarios. However, we found diminishing returns beyond two or three shots for tasks involving diverse semantic requirements or spatial reasoning. This suggests that the utility of additional examples is context-dependent and saturates relatively quickly in certain domains.


\begin{figure*}[t]
    \centering
    \includegraphics[width=\linewidth]{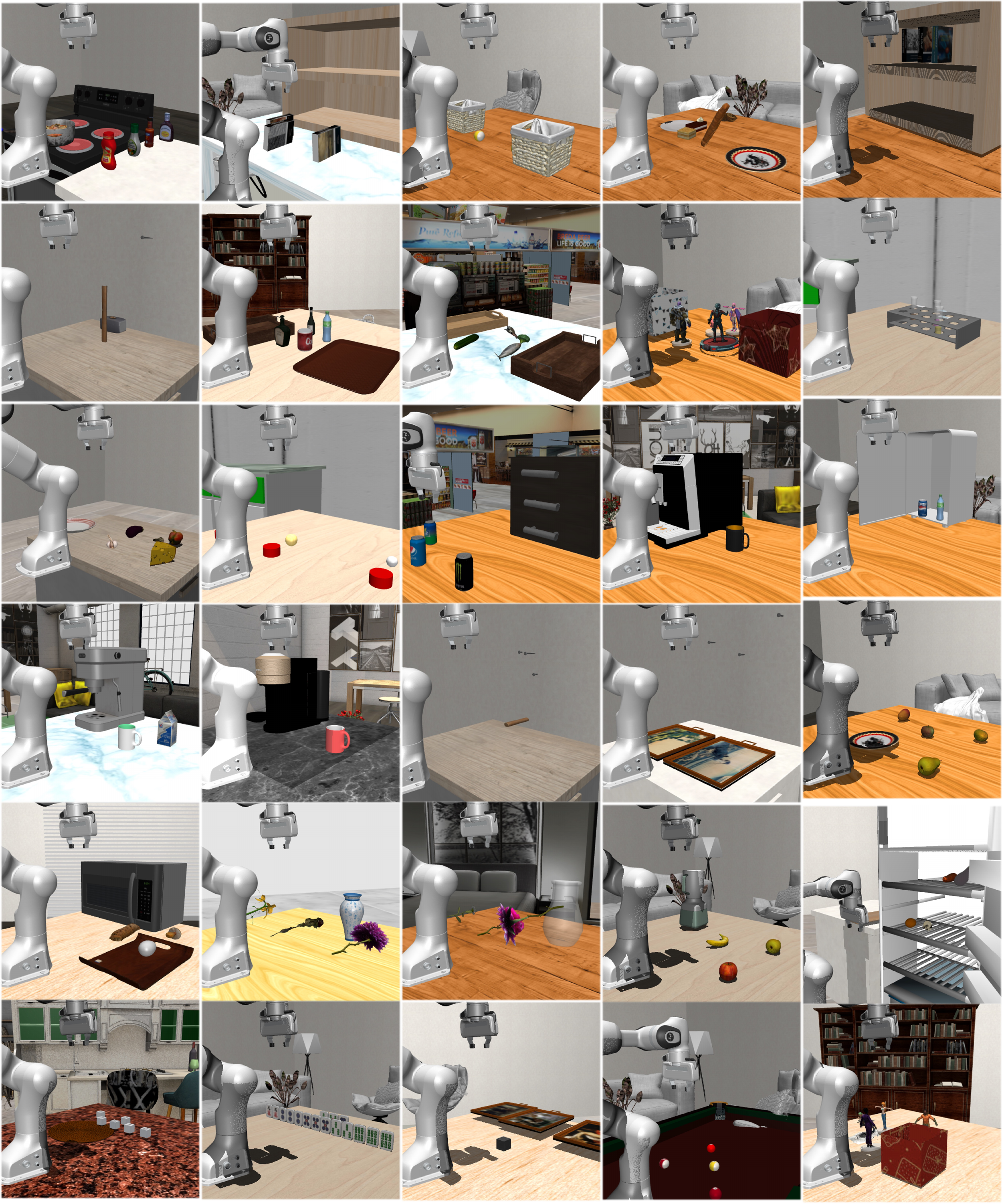}
    \caption{Part of tasks in VLABench.}
    \label{fig:all_tasks}
\end{figure*}

\onecolumn
\renewcommand{\arraystretch}{1.2}
\begin{longtable}{p{3cm} p{1.5cm} p{2cm} p{2.1cm} p{6cm}}
    \hline
    \textbf{Task} & \textbf{Type} & \textbf{Ability\newline Dimension} & \textbf{Skill\newline Involved} & \textbf{Description} \\ 
    \hline
    Select Fruit & Primitive & M\&T & Pick\&place & Pick the specific fruit into a specific receptacle, such as ``put the strawberry into the basket". \\ 
    \arrayrulecolor{lightgray}\hline
    Select Drink & Primitive & M\&T & Pick\&place, Pull & Get the specific drink from a particular receptacle, such as ``pick the cola out of the fridge". \\
    \arrayrulecolor{lightgray}\hline
    Select Toy & Primitive & M\&T & Pick\&place & Put the specific toy into a specific receptacle, such as ``Select Ironman from the toys and wrap it in the gift box". \\
    \arrayrulecolor{lightgray}\hline
    Select Book & Primitive & M\&T & Pick\&place, Pull & Take a particular book from the receptacle, such as ``Take \textit{Pride and Prejudice} from the bookshelf". \\
    \arrayrulecolor{lightgray}\hline
    Select Ingredient & Primitive & M\&T & Pick\&place & Get the specific ingredient from the particular receptacle, such as ``Take the bell pepper from the fridge and place it on the tray". \\
    \arrayrulecolor{lightgray}\hline
    Insert Flower & Primitive & M\&T & Pick\&place, Insert & Insert the specific flower into the container, such as ``Insert the rose into a vase."\\
    \arrayrulecolor{lightgray}\hline
    Add Condiment & Primitive & M\&T & Pick\&place, Pour & Add the specific condiment into the dish, such as ``Add some salt into the dish in the pot." \\
    \arrayrulecolor{lightgray}\hline
    Put Box on Famous Painting & Primitive & M\&T & Pick\&place & Place the geometric shape on the specified famous painting., such as ``Press the button before the painting \textit{The Starry Night}" \\
    \arrayrulecolor{lightgray}\hline
    Pick ChemistryTube & Primitive & M\&T & Pick\&place & Select specific solution tube based on the nametag, such as ``pick up the tube of CuCl2". \\
    \arrayrulecolor{lightgray}\hline
    Select Poker & Primitive & M\&T & Pick\&place & Select specific poker, such as ``Pick jack of red heart". \\
    \arrayrulecolor{lightgray}\hline
    Select Mahjong & Primitive & M\&T & Pick\&place & Select specific mahjong, such as ``Pick mahjong: 2 of Man and put it on the placemat". \\
    \arrayrulecolor{lightgray}\hline
    Select Billiards & Primitive & M\&T & Pick\&place & Select specific Billiards, such as ``Pick Black 8 and place it in any hole." \\
    \arrayrulecolor{lightgray}\hline
    Hammer Loose Nail & Primitive & M\&T & Pick\&place, Tool use & By comparing the lengths of different nails, use a hammer to tighten the loose nail, such as ``Hammer the loose nail on the wall". \\
    \arrayrulecolor{black}\hline
    
    Select Fruit-Spatial & Primitive & M\&T, SP & Pick\&place & Pick the fruit in a specific place or a certain spatial relationship, such as ``put the nearest strawberry into the plate". \\
   \arrayrulecolor{lightgray}\hline
    Select Drink-Spatial & Primitive & M\&T, SP & Pick\&place & Get the drink from a specific place or a certain spatial relationship, such as ``pick the monster outside", while there is also a can of monster inside the fridge. \\
   \arrayrulecolor{lightgray}\hline
    Select Toy-Spatial & Primitive & M\&T, SP & Pick\&place & Take the toy from a specific place or a certain spatial relationship, such as ``Place the toy on Luffy's right-hand side into the box". \\
   \arrayrulecolor{lightgray}\hline
    Select Book-Spatial & Primitive & M\&T, SP & Pick\&place & Take the book on the specific position or a certain spatial relationship, such as ``Take out the book on the most left in the top layer". \\
   \arrayrulecolor{lightgray}\hline
    Select Ingredient-\newline Spatial & Primitive & M\&T, SP & Pick\&place & Get the ingredient on the specific position or in a certain spatial relationship, such as ``Place the ingredient on the bottom layer in the fridge onto the tray". \\
   \arrayrulecolor{lightgray}\hline
    Insert Flower-Spatial & Primitive & M\&T, SP & Pick\&place & Insert the flower on the specific position or in a certain spatial relationship, ``Place the flower on the far left into the vase.". \\
   \arrayrulecolor{lightgray}\hline
    Add Condiment-\newline Spatial & Primitive & M\&T, SP & Pick\&place & Add the condiment on the specific position or in a certain spatial relationship, such as ``Add the furthest spice to the dish".\\
   \arrayrulecolor{lightgray}\hline
    Hang Picture & Primitive & M\&T, SP & Pick\&place, Hang & Hang the picture on the nail in the specified location, such as ``Hang the picture on the highest nail". \\
   \arrayrulecolor{lightgray}\hline
    Pick ChemistryTube-\newline Spatial. & Primitive & M\&T, SP & Pick\&Place & Take out the chemistry solution tube on the specific position or in a certain spatial relationship, such as ``Take the tube in the first row, the second column before you". \\
   \arrayrulecolor{lightgray}\hline
    Select Poker-Spatial & Primitive & M\&T, SP & Pick\&Place & Select the poker on the specific position or in a certain spatial relationship, such as ``Pick the second  poker from left to right".\\
   \arrayrulecolor{lightgray}\hline
    Select Mahjong-\newline Spatial & Primitive & M\&T, SP & Pick\&Place & Select the mahjong on the specific position or in a certain spatial relationship, such as ``Pick the mahjong on the right of six of sou". \\
   \arrayrulecolor{lightgray}\hline
    Put Billiards \newline in Pocket & Primitive & M\&T, SP & Pick\&Place & Place the billiard ball into the specified pocket, such as ``Place the 8-ball into the pocket in the right front". \\
   \arrayrulecolor{black}\hline

    Select Fruit \newline with Common Sense & Primitive & M\&T, C\&W & Pick\&Place  & Select fruit with specific characteristics, including nutritional characteristics, common uses, whether they grow in clusters, easy to peel, etc. Example: ``Put the fruit with the most vitamin C into the basket" from among \textbf{orange}, banana, and apple.\\
    \arrayrulecolor{lightgray}\hline
    Select Drink \newline with Common Sense & Primitive & M\&T, C\&W & Pick\&Place  & Select a drink with some specific characteristics including types of beverages, functions of the beverages, flavors of the beverages, etc. Example: ``Get a can of energy drink from the fridge" from among cola, apple juice, and \textbf{redbull}. \\
    \arrayrulecolor{lightgray}\hline
    Select Toy \newline with Common Sense & Primitive & M\&T, C\&W & Pick\&Place & Select the toy with some specific characteristics including the associated IP, character personality, character background, etc. Example: ``Put the toy from the Marvel series to the giftbox" from among \textbf{Hulk}, Batman, and Mickey. \\
    \arrayrulecolor{lightgray}\hline
    Select Book \newline with Common Sense & Primitive & M\&T, C\&W & Pick\&Place & Select the book with some specific characteristics including the type of the book, the content of the book, the main message it conveys, etc. Example: ``Get the book about computer science" from among \textit{Steve Jobs}, \textit{\textbf{3D Computer Vision}} and \textit{War and Peace}. \\ 
    \arrayrulecolor{lightgray}\hline
    Select Ingredient \newline with Common Sense & Primitive & M\&T, C\&W & Pick\&Place & Select the ingredient with some specific characteristics, such as ``Pick an ingredient full of protein from the fridge and put it on the tray" from among \textbf{egg}, tomato and bell pepper. \\
    \arrayrulecolor{lightgray}\hline
    Insert Flower \newline with Common Sense & Primitive & M\&T, C\&W & Pick\&Place & Insert the flower with some specific characteristics into vase including the flower language, the symbolic qualities of the flower, appropriate occasions for giving flowers, etc. Example:``Insert the flower suitable for Valentine's Day into the vase" from \textbf{rose}, sunflower, and tulip. \\
    \arrayrulecolor{lightgray}\hline
    Insert Bloomed \newline Flower & Primitive & M\&T, C\&W & Pick\&Place & An intelligent agent should possess awareness: flowers should be arranged with blooming ones, not with those that are already withered. Example: ``Insert a proper flower into the vase" from among wilted rose, wilted daisy, and \textbf{sunflower}. \\
    \arrayrulecolor{lightgray}\hline
    Add Condiments \newline with Common Sense & Primitive & M\&T, C\&W & Pick\&Place & Add the condiment with some specific characteristics including distinctive flavor, seasoning role, suitability for various dishes, and etc. Example: ``Add the condiment that makes the dish taste more salty" from among \textbf{salt}, ketchup, and salad dressing.\\
    \arrayrulecolor{lightgray}\hline
    Select Painting \newline with Common Sense & Primitive & M\&T, C\&W & Pick\&Place & Press the button before the painting with specific styles or contents. Example: ``Choose the painting in the style of rococo" among from paintings of \textit{\textbf{La Liseuse}}, \textit{The Stary Night}, and \textit{Golden Autumn}. \\
    \arrayrulecolor{lightgray}\hline   
    Pick Chemistry Tube \newline with Common Sense & Primitive & M\&T, C\&W & Pick\&Place & Pick up the specific solution without the nametag and distinguish by the solution color. Example: ``Pick up the solution of CuSO4" from among the solution of \textbf{blue}, green, and yellow. \\
    \arrayrulecolor{lightgray}\hline
    Select nth \newline Largest Poker & Primitive & M\&T, C\&W & Pick\&Place & Choose the largest poker under the rule of the specific poker games. Example: ``Pick the largest poker in single under the rule of Texas Holdem" from among \textbf{Ace of Spades}, Three of Hearts, and Queen of Clubs. \\
    \arrayrulecolor{lightgray}\hline
    Select Unique \newline Mahjong & Primitive & M\&T, C\&W & Pick\&Place & Choose the mahjong with the unique type. Example:``Pick the unique type of mahjong" among from \textbf{East}, One of Man, Nine of Man. \\
    \arrayrulecolor{lightgray}\hline
    Select Billiards \newline with Common Sense & Primitive & M\&T, C\&W & Pick\&Place & Select a specific billiard game under particular rules with a specific score. Example: ``Place the two-point ball from a snooker match into any pocket" from among green ball, \textbf{yellow ball}, and red ball. \\
    \arrayrulecolor{black}\hline    
    
    Select Fruit-\newline Semantic & Primitive & M\&T, SEM & Pick\&Place & The user expresses implicit needs for a certain fruit during a semantically rich conversation or context, such as: ``Today, I suddenly feel like doing some baking and plan to make a strawberry cake! Could you help me prepare the fruits I’ll need?"\\
    \arrayrulecolor{lightgray}\hline
    Select Drink-\newline Semantic & Primitive & M\&T, SEM & Pick\&Place,\newline Pull & The user expresses implicit needs for a certain drink during a semantically rich conversation or context, such as: ``I just worked out at the gym for a long time, and now I'm a bit dehydrated. Could you help me grab a bottle of electrolyte drink from the fridge?"\\
    \arrayrulecolor{lightgray}\hline
    Select Toy-\newline Semantic & Primitive & M\&T, SEM & Pick\&Place & The user expresses implicit needs for a specific toy during a semantically rich conversation or context, such as ``I've loved Disney since I was a kid, especially the Toy Story series! I want to place Buzz Lightyear on the top layer of the shelf, but I can't reach it! " \\
    \arrayrulecolor{lightgray}\hline
    Select Book-\newline Semantic & Primitive & M\&T, SEM & Pick\&Place, \newline Pull & The user expresses implicit needs for a specific book during a semantically rich conversation or context, such as ``I'm getting ready to review for my final Python exam. Could you help me prepare the textbook?"\\
    \arrayrulecolor{lightgray}\hline
    Select Ingredient-\newline Semantic & Primitive & M\&T, SEM & Pick\&Place & The user expresses implicit needs for a specific ingredient during a semantically rich conversation or context, such as ``I'm keeping fit so I want to eat something full of protein. I want a steak as my lunch and could you get one for me?" \\
    \arrayrulecolor{lightgray}\hline
    Insert Flower-\newline Semantic & Primitive & M\&T, SEM & Pick\&Place & The user expresses implicit needs for a specific flower during a semantically rich conversation or context, such as ``Today is Teacher's Day, and Ms. Lisa has always been kind to me. I want to give her a bouquet of carnations. Could you help me place them in the vase on her desk?" \\
    \arrayrulecolor{lightgray}\hline
    Add Condiment-\newline Semantic & Primitive & M\&T, SEM & Pick\&Place,\newline Pour & The user expresses implicit needs for a specific condiment during a semantically rich conversation or context, such as ``I'm making tomato-braised beef brisket, but the tomato flavor doesn't seem strong enough. Could you help me add some tomato paste? Thanks!" \\
    \arrayrulecolor{lightgray}\hline
    Select Painting-\newline Semantic & Primitive & M\&T, SEM & Press & The user expresses implicit needs for a specific solution during a semantically rich conversation or context, such as: ``I am a student who has just started learning painting, and I’m not very good at distinguishing between different styles of painting. Could you help me identify which of these three paintings is in the realist style?"\\ 
    \arrayrulecolor{lightgray}\hline
    Select ChemistryTube-\newline Semantic & Primitive & M\&T, SEM & Pick\&Place & The user expresses implicit needs for a specific solution during a semantically rich conversation or context, such as: ``I'm going to demonstrate an acid-base neutralization experiment today, but I’m missing an acid-base indicator. Could you help me grab the phenolphthalein solution?" \\
    \arrayrulecolor{lightgray}\hline
    Simple Poker \newline Play & Primitive & M\&T, SEM & Pick\&Place & The agent plays the poker that should be played on behalf of the player during the semantically rich interaction. Example: ``We're playing Landlord, and the player before me just played a 10. Now it's our turn. Please play a 2 for me."\\
    \arrayrulecolor{lightgray}\hline
    Simple Mahjong \newline Play & Primitive & M\&T, SEM & Pick\&Place & The agent plays the Mahjong that should be played on behalf of the player during the semantically rich interaction. Example: ``It's hard to win with the 'Wan' character tiles left. Go ahead and discard the '1 Wan'." \\
    \arrayrulecolor{lightgray}\hline
    Simple Snooker \newline Play & Primitive & M\&T, SEM & Pick\&Place & The agent picks the billiard that should be played on behalf of the player during the semantically rich interaction. ``We're playing a simple game of snooker. Now, let's pot the yellow ball into the pocket."\\
    \arrayrulecolor{black}\hline
    
    Friction QA & Primitive & M\&T, PHY & Press & Using the relevant physics knowledge of sliding friction and rolling friction, determine the rolling speed of different shaped and material objects on a slope. Example: ``Press the button before the object that falls the fastest down the slope." \\
    \arrayrulecolor{lightgray}\hline
    Density QA & Primitive & M\&T, PHY & Press & Visually judge the material of an object and determine the relative density of objects made from different materials. Example: ``Press the button before the object can float on water." \\
    \arrayrulecolor{lightgray}\hline
    Magnetism QA & Primitive & M\&T, PHY & Press & Visually identify the material of an object and determine whether objects made from different materials are magnetic. Example: ``Press the button before the object that is not magnetic." \\
    \arrayrulecolor{lightgray}\hline
    Weight QA & Primitive & M\&T, PHY & Press & Visually identify the material of an object, and combine the material density and shape (in the actual setup, this includes cubes of different shapes, along with their corresponding inscribed spheres and circumscribed spheres) to make a comprehensive judgment of the object's mass. Example: ``Press the button before the object with the smallest weight." \\
    \arrayrulecolor{lightgray}\hline
    Thermal Expansion \newline QA & Primitive & M\&T, PHY & Press & Visually identify the material of an object and determine the thermal expansion properties of objects made from different materials. Example: ``Press the button before the object with a medium thermal expansion coefficient." \\
    \arrayrulecolor{lightgray}\hline
    Speed of Sound QA & Primitive & M\&T, PHY & Press & Visually identify the material of an object and determine the sound propagation speed in objects made from different materials. Example: ``Press the button before the object that sound propagates fastest in." \\
    \arrayrulecolor{lightgray}\hline
    Specular Reflection \newline QA & Primitive & M\&T, PHY & Press & Judge based on visual information whether different objects exhibit specular reflection and make a selection. Example: ``Press the button before the object that can reflect the image of others."\\
    \arrayrulecolor{lightgray}\hline
    Drag Force QA & Primitive & M\&T, PHY & Press & Determine the object's free fall speed based on its shape, texture, and material. This involves physical theories such as air resistance, the Kármán vortex street effect, and others. Example: ``Press the button before the object falls slowest in the air" from among \textbf{golf}, basketball, football. \\
    \arrayrulecolor{lightgray}\hline
    Basic Seesaw Usage & Primitive & M\&T, PHY & Pick\&place,\newline Tool use & Using the principle of leverage, place a heavy object on one side of the seesaw to lift the other side. Example: ``Make the other side of the seesaw lift". \\
    \arrayrulecolor{lightgray}\hline
    Strike Billiards & Primitive & M\&T, PHY & Pick\&place,\newline Tool use & Use the laws of collision to perform a simple strike. Example: ``Use the cue stick to strike the white ball, aiming to make it hit other colored balls". \\
    \arrayrulecolor{black}\hline
    
    Take Chemistry\newline Experiment & Composite  & M\&T, SP,\newline SEM, C\&W, \newline L\&R & Pick\&place, Insert,\newline Pour & The agent should first use the user's request for the desired chemical product, \textbf{combine it with visual observation and common knowledge for logical reasoning}, and determine the chemical \textbf{solutions involved in the reaction}. After identifying the appropriate solutions using the name tag, the agent should select the solutions and mix them into the flask. Example: ``I would like to obtain AgCl precipitation in the flask. Please carry out this experiment." \\
    \arrayrulecolor{lightgray}\hline
    Find Unseen Object & Composite & M\&T, SP,\newline L\&R & Open\&close drawer, Pick\&Place, \newline Explore & The target object is \textbf{not directly visible}, requiring the agent to open multiple drawers and eventually find the target object. Example: ``Find a snack in the drawer for me".\\
    \arrayrulecolor{lightgray}\hline
    Find Unseen Object \newline
    without Telling Find & Composite & M\&T, SP,\newline SEM, C\&W, \newline L\&R & Open\&close drawer, Pick\&Place, \newline Explore & The other settings are the same as for Find Unseen Object, but the requirements are implicitly conveyed through semantically rich dialogue. The agent needs to be aware of the need for \textbf{exploration and search on its own}. Example: ``I'm a bit hungry, could you get me something to eat?" \\
    \arrayrulecolor{lightgray}\hline
    Make Juice\newline with Juicer & Composite & M\&T, SEM, \newline L\&R & Pick\&place,\newline Tool use,\newline Press  & Select the appropriate fruits based on semantically rich user instructions, place them into a container, and correctly use the juicer. Example: ``It's so hot today! I feel like having a freshly squeezed kiwi and strawberry juice right now."\\
    \arrayrulecolor{lightgray}\hline
    Find Fruit to\newline Make Juice & Composite  & M\&T, C\&W, \newline SEM, L\&R & Pick\&place,\newline Tool use,\newline Press, \newline Explore & The fruits are \textbf{not directly available and visible} to the agent because the fruits are stored in a closed fridge or a cabinet. The agents should find the proper fruit first. The example is the same as above.\\
    \arrayrulecolor{lightgray}\hline
    Plug-in Power Cord\newline to Make Juice & Composite & M\&T, C\&W, \newline SEM, L\&R & Pick\&place,\newline Tool use,\newline Press, \newline Explore,\newline Insert & The other basic settings remain the same as above. However, the juicer's \textbf{power cord is not plugged in}. The agent needs to first observe this and, using common sense, plug in the power cord to supply power. The example is the same as above.\\
    \arrayrulecolor{lightgray}\hline
    Take out Cool Drink & Composite  & M\&T, SP, \newline C\&W, SEM,\newline  L\&R & Open\&close door,\newline Pick\&place & Obtain user requirements through semantically rich interaction: the user wants a cold drink. Given the observation of the same target \textbf{drink on the desk as disturbance}, the agent should use \textbf{common sense to determine that the drink from the fridge} should be chosen. Example: ``The weather is so hot! I feel like having a cold soda."\\
    \arrayrulecolor{lightgray}\hline
    No Drink in Fridge \newline \& Refrigerate Drink & Composite  & M\&T, SP, \newline C\&W, SEM,\newline  L\&R & Open\&close door,\newline Pick\&place, \newline Explore & The task is set the same as above. However, after the agent opens the fridge door, it finds that the target object is not there. The agent needs to realize that it should first \textbf{refrigerate the room-temperature target drink}. \\
    \arrayrulecolor{lightgray}\hline
    Wrap Proper Toy \newline as Gift & Composite  & M\&T, C\&W,\newline SEM, L\&R & Open\&close door,\newline Pick\&place & Choose a suitable toy for kids as a gift from product shelf during the semantic interaction with the user. Then wrap in as a gift. Example: ``My son is a superhero fun, but I don't know that much. Could you wrap a gift for him?" \\
    \arrayrulecolor{lightgray}\hline
    Rearrange Books\newline by Year & Composite  & M\&T, C\&W, \newline SP, L\&R & Pick\&place & Identify the book title and use \textbf{world knowledge to determine the publication period}. Then rearrange them. Example: ``Rearrange the book by published year order in the top layer of the shelf, the far left is the earliest one." \\
    \arrayrulecolor{lightgray}\hline
    Rearrange Books \newline by Author Name & Composite  & M\&T, C\&W, \newline SP, L\&R & Pick\&place & Identify the book title and use \textbf{world knowledge to determine the author name}. Then rearrange them. Example: ``Rearrange the book by their author names, the far right starts with the largest word."\\
    \arrayrulecolor{lightgray}\hline
    Classify the Books & Composite  & M\&T, C\&W, \newline SP, L\&R & Pick\&place & Identify the book titles and categorize the books based on their genre or content. The agent needs to \textbf{infer the classification criteria} on its own and correctly divide the books into two layers. Example: ``Divide the books into two classes, one class on the top layer while another on the bottom."\\
    \arrayrulecolor{lightgray}\hline
    Cook Dishes \newline Following Menu& Composite  & M\&T, C\&W, \newline SEM, L\&R & Pick\&place & \textbf{Multi-turn pick and place the correct ingredients} for a dish whose menu is offered by semantic instructions. Example: ``I'm about to cook a dish of tomato-fried eggs, prepare ingredients in the tray." \\
    \arrayrulecolor{lightgray}\hline
    Store Proper Food & Composite  & M\&T, C\&W, \newline SEM, L\&R & Open\&close door, Pick\&place & Store the ingredients or fruits into the fridge and do not put the disturbance including snacks into the fridge. Example: ``I left some food on the table in the last meal, store them properly please." \\
    \arrayrulecolor{lightgray}\hline
    Heat Food with\newline Microwave & Composite  & M\&T, C\&W, \newline SEM, L\&R & Open\&close door, Pick\&place, Press & \textbf{Extract implicit goals} from semantically rich interactions: heating food. Use common sense to choose the proper food, such as a hot dog, with the microwave, \textbf{while avoiding heating canned food or raw ingredients}. Finally, correctly use the microwave. Example: ``I just finished class, and now my stomach is growling. Could you heat up some food for me to have a quick bite?"\\
    \arrayrulecolor{lightgray}\hline
    Plug-in Power Cord\newline to Heat Food & Composite  & M\&T, C\&W, \newline SEM, L\&R & Open\&close door, Pick\&place, Press,\newline Insert & The other experimental settings remain the same as above. The agent must first have the \textbf{common sense to plug in the power source} for the device to operate. The example is the same as above.\\
    \arrayrulecolor{lightgray}\hline
    Replace Wilted\newline Flower and Drop & Composite  & M\&T, C\&W, \newline SEM, L\&R & Pick\&place, \newline Insert & Based on the semantically rich user request and using common sense, determine the target flower. Discard the wilted flower in the vase, and then insert the new flower. Example: ``It's Valentine's Day today, replace the flower in the vase."   \\
    \arrayrulecolor{lightgray}\hline
    Find Condiment\newline and Add to Dish & Composite  & M\&T, C\&W, \newline SEM, SP,\newline L\&R & Open\&close drawer, Pick\&Place, Explore,\newline Pour & All the condiments are stored in the cabinet and the agent should \textbf{proactively find them first} and add proper condiment into the dish. Example: ``The spiciness of this dish isn't quite enough. Could you add some more seasoning to make it tastier?"\\
    \arrayrulecolor{lightgray}\hline
    Hammer Nail \newline \&Hang Picture & Composite  & M\&T, C\&W, \newline SEM, L\&R & Pick\&place, Tool use, \newline Hang & The agent needs to observe and determine if the nail is loose, and then use a hammer to tighten the nail. After that, the agent should hang the appropriate picture on the wall. Example:  ``Hang 'the Stary Night' on the wall steadily." \\
    \arrayrulecolor{lightgray}\hline
    Assemble Hammer\newline \&Hammer Nail & Composite  & M\&T, C\&W, \newline L\&R & Pick\&place, Insert, \newline Tool Use & The agent needs to observe and \textbf{reason that the task cannot be completed with the current conditions}. It must first assemble the hammer handle and the hammerhead precisely before proceeding. Example: ``Hammer the loose nail." \\ 
    \arrayrulecolor{lightgray}\hline
    Rearrange Chemistry Tube & Composite  & M\&T, SP,\newline C\&W, L\&R & Pick\&place,\newline Insert & Rearrange the multiple tubes by the \textbf{corresponding relationships between color and name tag}, the result of utilizing common sense and reasoning ability. Example: ``Rearrange the solution tubes." \\ 
    \arrayrulecolor{lightgray}\hline
    Texas Holdem Play & Composite  & M\&T, C\&W,\newline SEM, L\&R & Pick\&Place & \textbf{Deduce the strongest Texas Hold'em hand based on the common game rules and visual information}. Then take multi-step pick\&place. Example: ``We are playing Texas Hold'em, place your strongest hand combination on the placemat." \\
    \arrayrulecolor{lightgray}\hline
    Flip Facing-downs\newline \&Play Texas Holdem & Composite  & M\&T, C\&W,\newline SEM, L\&R & Pick\&Place,\newline Twist,\newline Explore & Based on the previous task, some of the cards are face down. The agent needs to have an \textbf{exploration mindset and actively retrieve all the observational information}, then make the correct judgment. The example is the same as above.\\
    \arrayrulecolor{lightgray}\hline
    Play Mahjong & Composite  & M\&T, C\&W,\newline SEM, L\&R & Pick\&Place & The agent \textbf{makes decisions based on world knowledge of Mahjong rules combined with visual information}. It discards an unnecessary tile and draws a necessary tile to win. Example: ``We seem to be close to winning the game. Take the right actions to help us win this round."\\
    \arrayrulecolor{lightgray}\hline
    Flip Facing-downs\newline \&Play Mahjong & Composite  & M\&T, C\&W,\newline SEM, L\&R & Pick\&Place,\newline Twist,\newline Explore & The agent needs to have an \textbf{exploration mindset and actively retrieve all the observational information}, then make the correct judgment. The example is the same as above.\\
    \arrayrulecolor{lightgray}\hline
    Leverage SeeSaw to\newline Grasp Target & Composite  & M\&T, C\&W,\newline SEM, PHY,\newline L\&R & Pick\&place, Explore, \newline Tool use & Using the lever principle, place one or more heavy objects on one side of the seesaw to lift the target object on the other side, initially unreachable. The challenge lies in the fact that \textbf{if the placed weights are insufficient, the agent will need to add additional weights}. Example:``I want to eat the pear in the glass container, but I can't get it out. Can you help me?" \\
    \arrayrulecolor{lightgray}\hline
     Find Weights to\newline Leverage SeeSaw & Composite  & M\&T, C\&W,\newline SEM, SP,\newline PHY, L\&R & Open\&close drawer, Pick\&place,\newline Explore,\newline Tool use & All the weights are stored in the cabinet and are not visible. The agent needs to \textbf{explore multiple drawers to find enough weights} before being able to properly use the seesaw. The example is the same as above.\\
     \arrayrulecolor{lightgray}\hline
    Get Black Coffee & Composite  & M\&T, C\&W, \newline SEM, L\&R & Pick\&place, \newline Tool use,\newline Press & The agent needs to derive the task goal from semantically rich interactions: to prepare a cup of \textbf{coffee without milk and sugar}. Then, it should correctly place the cup and operate the coffee machine. Example: ``I'm feeling sleepy right now. Could you get me a cup of coffee? An Americano will do." \\
    \arrayrulecolor{lightgray}\hline
    Get Sweet Coffee & Composite  & M\&T, C\&W, \newline SEM, L\&R & Pick\&place, \newline Tool use,\newline Press,\newline Pour & The agent needs to additionally infer firstly: the user prefers sweet coffee -> \textbf{the coffee needs to be prepared with sugar}. Example: ``Get me a cup of sweet coffee to clear my mind, thx!" \\
    \arrayrulecolor{lightgray}\hline
    Get Latte Coffee & Composite  & M\&T, C\&W, \newline SEM, L\&R & Pick\&place, \newline Tool use,\newline Press,\newline Pour & The agent needs to additionally infer firstly: \textbf{latte coffee is composed of black coffee and milk}. Example: ``A cup of latte please. Nice to meet you here, thank god!" \\
    \arrayrulecolor{lightgray}\hline
    Set Dining Table\newline by Menu& Composite  & M\&T, C\&W, \newline SEM, SP, \newline L\&R & Pick\&place & The agent needs to \textbf{infer the appropriate utensils based on semantic interactions and the type of cuisine}. For example, chopsticks for Chinese cuisine, a knife and fork for Western cuisine, and a spoon if soup is being served. Example: ``Today's main course is steak! Please help me set up the table."\\
    \arrayrulecolor{lightgray}\hline
    Set Dining Table\newline Left-Handed & Composite  & M\&T, C\&W, \newline SEM, SP, \newline L\&R & Pick\&place & The agent needs to first extract the key information from user interactions that the user is \textbf{left-handed}. Then, using common sense, it should \textbf{adjust the placement of the utensils}, such as switching from the original left-knife-right-fork arrangement to a left-fork-right-knife setup. Example: ``Tonight, we're having fried rice! Remember to get me a spoon. Oh, and don't forget that I'm left-handed."\\
    \arrayrulecolor{lightgray}\hline
    Play Snooker & Composite  & M\&T, C\&W, \newline SEM, L\&R & Pick\&place & Put the ball into the hole by snooker order: yellow, green, brown, blue, pink, black. The agent needs to make the \textbf{correct sequence of decisions based on snooker rules in world knowledge}. Example: ``Put the colored billiards into holes by score order in a snooker match."\\
    \arrayrulecolor{lightgray}\hline
    Cluster Toy & Composite  & M\&T, C\&W,\newline SP, L\&R & Pick\&place & Based on common sense, world knowledge, and visual information, cluster the toys according to their \textbf{associated IPs, character types}, and other attributes. Example: ``Cluster the toys into two classes." These toys are \textcolor{red}{Spiderman}, \textcolor{red}{Hawk Eye}, \textcolor{blue}{Nami}, \textcolor{blue}{Chopper}.  \\
    \arrayrulecolor{lightgray}\hline
    Classify Desserts & Composite  & M\&T, C\&W,\newline SP, L\&R & Pick\&place & Based on common sense, world knowledge, and visual information, categorize the desserts according to their types. Example: ``Classify the different desserts." These desserts are \textcolor{red}{strawberry donut}, \textcolor{red}{banana donut}, \textcolor{blue}{coco cupcake}, \textcolor{blue}{common cupcake}.\\
    \arrayrulecolor{lightgray}\hline
    Setup Study Table & Composite & M\&T, C\&W,\newline SP, L\&R & Pick\&place\newline Open laptop &Determine the task goal from semantically rich interactions: the user needs a specific book and to use the computer. The agent needs to use common sense to \textbf{infer the correct book} and place it on the desk, while also \textbf{turning on the computer}. Example: ``I have a Python practical exam the day after tomorrow, and I'm planning to review later. Could you help me set up my desk?"\\
    \arrayrulecolor{lightgray}\hline
    Organize Study Table & Composite & M\&T, C\&W,\newline SP, L\&R & Pick\&place\newline Close laptop & Determine the task goal by observing the desk and combining user interactions: organize the desk. This requires completing subtasks in sequence, including \textbf{arranging the books} and closing the laptop. Example: ``That's all for today. Please help me tidy up the desk. Thanks!"\\
    \arrayrulecolor{lightgray}\hline
    Math Game & Composite  & M\&T, C\&W,\newline SEM, L\&R & Pick\&place & Based on the \textbf{math problem provided by the user}, use \textbf{logical reasoning to find the answer} and display it by arranging number blocks to form the solution. Example: ``Let's play a math game, show me the answer by number blocks. The question is: `Toulouse has twice as many sheep as Charleston. Charleston has 4 times as many sheep as Seattle. How many sheep do Toulouse, Charleston, and Seattle have together if Seattle has 20 sheep?'" This math question is from the GSM8K dataset \cite{cobbe2021training}.\\
    \arrayrulecolor{lightgray}\hline
    Art Game & Composite & M\&T, C\&W,\newline SEM, SP,\newline PHY, L\&R & Pick\&place & Place the geometric object with a \textbf{specific physical property} onto the painting that aligns with the user's \textbf{hinted content or style}. Example: ``Let's play a game of 'Simon Says'! Place the geometric object with a specific physical property onto the painting that aligns with the user's hinted content or style.".\\
    \arrayrulecolor{lightgray}\hline
    Cluster Beverage & Composite & M\&T, C\&W,\newline SP, L\&R & Pick\&place & Based on common sense, world knowledge, and visual information, cluster the drinks according to their types. Example: ``Cluster the beverages into two types." These beverages are \textcolor{red}{mango juice}, \textcolor{red}{milk}, \textcolor{blue}{Vodka}, \textcolor{blue}{Champagne}.\\
    \arrayrulecolor{black}\hline
    \caption{Task List. Include the name, type, ability required, and detailed description of all the tasks.}
    \label{table:task_description}
\end{longtable}
\twocolumn


\end{document}